\documentclass[lettersize,journal]{IEEEtran}
\usepackage{etoolbox}
\makeatletter
\@ifundefined{color@begingroup}%
{\let\color@begingroup\relax
\let\color@endgroup\relax}{}%
\def\fix@ieeecolor@hbox#1{%
\hbox{\color@begingroup#1\color@endgroup}}
\patchcmd\@makecaption{\hbox}{\fix@ieeecolor@hbox}{}{\FAILED}
\patchcmd\@makecaption{\hbox}{\fix@ieeecolor@hbox}{}{\FAILED}
\usepackage{amsmath,amsfonts}
\usepackage{algorithmic}
\usepackage{algorithm}
\usepackage{amsmath}
\usepackage{subfigure}  
\usepackage{array}
\usepackage[table,xcdraw]{xcolor}
\usepackage{multirow}
\usepackage[colorlinks, linkcolor=blue, citecolor=blue, urlcolor=blue, filecolor=blue]{hyperref}
\usepackage{mathrsfs}
\usepackage{textcomp}
\usepackage{stfloats}
\usepackage{url}
\usepackage{verbatim}
\usepackage{graphicx}
\usepackage{cite}
\usepackage{orcidlink}
\usepackage{cleveref}
\usepackage{subcaption}
\usepackage{float}

\begin{document}

\title{You Only Acquire Sparse-channel (YOAS): A Unified Framework for Dense-channel EEG Generation}

\author{Hongyu Chen\textsuperscript{\large\orcidlink{0009-0004-2934-3135}},  Weiming Zeng\textsuperscript{\large\orcidlink{0000-0002-9035-8078}}, \IEEEmembership{Senior Member, IEEE}, Luhui Cai\textsuperscript{\large\orcidlink{0009-0002-3150-6664}}, 
Lei Wang\textsuperscript{\large\orcidlink{0000-0003-0111-4328}}, 
Jia Lu, Yueyang Li\textsuperscript{\large\orcidlink{0009-0008-5310-124X}}, Hongjie Yan\textsuperscript{\large\orcidlink{0009-0000-2553-2183}},
Wai Ting Siok\textsuperscript{\large\orcidlink{0000-0002-2154-5996}}
and Nizhuan Wang\textsuperscript{\large\orcidlink{0000-0002-9701-2918}}

\thanks{This work was supported by the National Natural Science Foundation of China [grant number 31870979]. (Corresponding author: Weiming Zeng, and Nizhuan Wang)}
\thanks{ Hongyu Chen, Weiming Zeng, Luhui Cai,  Lei Wang, Jia Lu, Yueyang Li are with the Laboratory of Digital Image and Intelligent Computation, Shanghai Maritime University, Shanghai 201306, China (e-mail: hongychen676@gmail.com, zengwm86@163.com, clh0x123@126.com, sayhiwl@163.com, lujia@shmtu.edu.cn, lyy20010615@163.com).}
\thanks{Hongjie Yan is with Department of Neurology, Affiliated Lianyungang Hospital of Xuzhou Medical University, Lianyungang 222002, China (email: yanhjns@gmail.com).}
\thanks{ Wai Ting Siok and Nizhuan Wang are with Department of Chinese and Bilingual Studies, The Hong Kong Polytechnic University, Hong Kong, SAR, China (e-mail: wai-ting.siok@polyu.edu.hk, wangnizhuan1120@gmail.com).}


\thanks{Manuscript received XX, XXXX; revised XXXX XX, XXXX.}}

\markboth{Journal of \LaTeX\ Class Files,~Vol.~XX, No.~X, June~2024}%
{Shell \MakeLowercase{\textit{et al.}}: A Sample Article Using IEEEtran.cls for IEEE Journals}


\maketitle
 
\begin{abstract}
High-precision acquisition of dense-channel electroencephalogram (EEG) signals is often impeded by the costliness and lack of good portability. In contrast, generating dense-channel EEG signals from sparse channels shows promise and economic viability. However, sparse-channel EEG poses challenges such as reduced spatial resolution, information loss, signal mixing, and heightened susceptibility to noise and interference. To address these challenges, we first theoretically formulate the dense-channel EEG generation problem as by optimizing a set of cross-channel EEG signal generation problems. Then, we propose the YOAS framework for generating dense-channel data from sparse-channel EEG signals. The YOAS totally consists of four sequential stages: Data Preparation, Data Preprocessing, Biased-EEG Generation, and Synthetic EEG Generation. Data Preparation and Preprocessing carefully consider the distribution of EEG electrodes and low signal-to-noise ratio problem of EEG signals. Biased-EEG Generation includes sub-modules of BiasEEGGanFormer and BiasEEGDiffFormer, which facilitate long-term feature extraction with attention and generate signals by combining electrode position alignment with diffusion model, respectively. Synthetic EEG Generation synthesizes the final signals, employing a deduction paradigm for multi-channel EEG generation. Extensive experiments confirmed YOAS's feasibility, efficiency, and theoretical validity, even remarkably enhancing data discernibility. This breakthrough in dense-channel EEG signal generation from sparse-channel data opens new avenues for exploration in EEG signal processing and application. 
\end{abstract}
\begin{IEEEkeywords}
Sparse-channel, Dense-channel, EEG signal generation, GAN, Diffusion model, BCI
\end{IEEEkeywords}

\section{Introduction}
\label{sec:1}
\IEEEPARstart{E}{lectroencephalogram} (EEG), crucial for neuroscience and clinical diagnosis, captures neuronal electrical activity from the cerebral cortex\cite{cohen2017does}. In academical research, EEG offers new insights into brain function, elucidating cognitive processes, sleep patterns, neural mechanisms of brain disorders, etc \cite{abiri2019comprehensive}, \cite{su2024extracting}, \cite{li2021effective}, \cite{shrestha2023color}, \cite{wang2023ssgcnet}, \cite{li2024eeg}. In clinical settings, EEG aids in diagnosing epilepsy \cite{valipour2024diagnosing}, \cite{sadoun2023seizure}, sleep disorders \cite{hartmann2023subject}, \cite{udeshika2024insomnia}, \cite{supratak2020tinysleepnet}, and informs surgical decisions \cite{gascoigne2023incomplete}, \cite{sumsky2018decision}. It also can assesse treatment efficacy \cite{goerttler2024balancing} and contribute to advancements in brain-computer interface (BCI) technology \cite{pan2024lateral}, \cite{bai2023dreamdiffusion}, \cite{marcel2007person}, \cite{hu2024brain}, \cite{Lawhern_2018}, enhancing rehabilitation and quality of life for individuals with disabilities.

High-precision dense-channel EEG acquisition is essential across diverse applications, yet the cost and portability limitations of dense-channel EEG devices pose significant challenges. A raised question is that is it feasible to develop a framework capable of generating multi-channel EEG signals from a limited number of channels?

Neural activity is influenced by factors such as neuron types, synaptic connections, and neural circuits \cite{Ajra_2023}. EEG signals recording neural activity are high-dimensional and dynamic, necessitating the capture of intricate temporal correlations and spatial distributions \cite{hector2021distributed}, \cite{duda2023time}. Thus, generating EEG signals presents challenges in simulating the spatiotemporal properties of each channel. Moreover, generated EEG signals exhibit uncertainty and variability due to physiological and environmental factors, posing limitations to model training and performance evaluation \cite{he2017generative}, \cite{lee2021contextual}. In summary, the generation of dense-channel EEG signals faces the following challenges:

Firstly, as each electrode corresponds to specific brain areas, harmonizing signal characteristics from electrodes at different locations under the same mental condition presents challenges. This is due to that the distinct characteristics of EEG signals vary among individuals across different inner states \cite{girard2023enhanced}, \cite{peng2022joint}, and the electrode signals exhibit unique features at various positions \cite{martinez2019automatic}, \cite{liu2021positional}. 

Secondly, EEG signal generation for whole brain based on few channels is exceptionally challenging. Initially, the low spatial resolution of EEG makes pinpointing their sources quite difficult. Further, within the International 10-20 system, the channel signals in the neighboring areas may have weak correlation, while those from distant areas may exhibit strong correlation. Additionally, the previous studies \cite{aznan2019simulating} have only focused on generating EEG data as augmentation. 

Thirdly, the generative models of EEG signals pose the challenges including training instability, inefficiency, and susceptibility to signal noise. Despite the advancement of generative models such as Variational Auto Encoder (VAE) \cite{Kingma2013AutoEncodingVB}, Generative Adversarial Network (GAN) series \cite{hartmann2018eeg}, and Diffusion models \cite{croitoru2023diffusion}, \cite{xia2024diffusion} in general domain, these issues persist for the specific EEG signal generation problem. 
\begin{figure*}
	\centering
	\includegraphics[width=\textwidth]{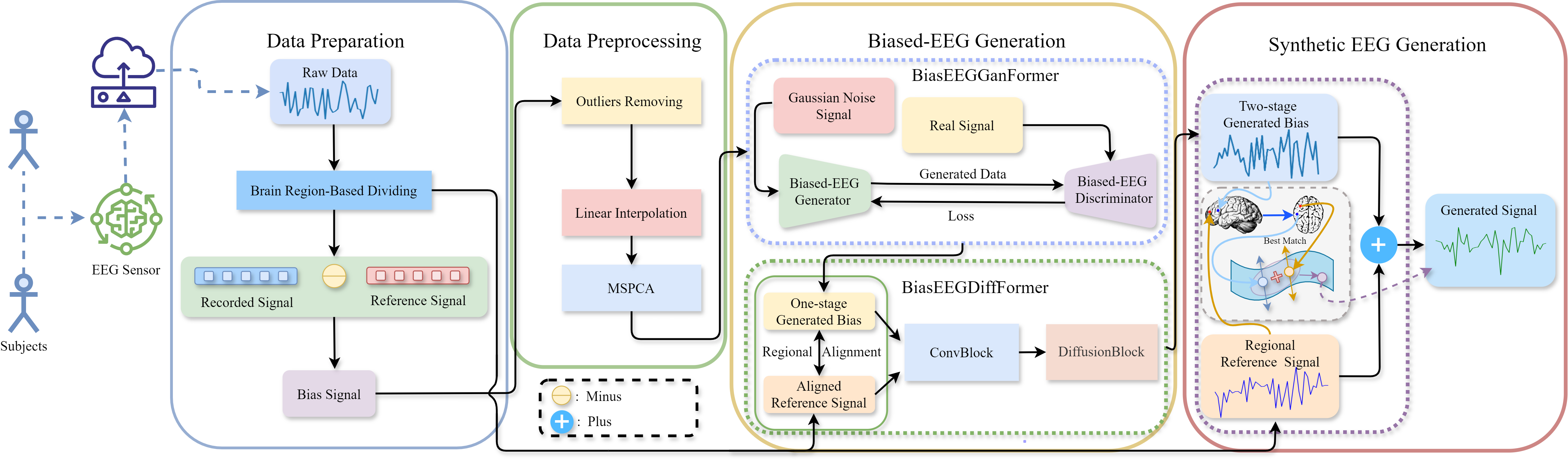}
	\caption{The YOAS framework. The Data Preparation module extracts Bias Signal from Raw Data through regional division, followed by Data Preprocessing to eliminate outliers. The Biased-EEG Generation module captures long-term features and aligns spatial positions to generate Two-stage Bias. Finally, the Synthetic EEG Generation module synthesizes the final signals.}
	\label{fig:fig1}
	\medskip
	\small 
\end{figure*}

To address the aforementioned challenges, we proposed a specialized model for dense-channel EEG generation problem, named YOAS comprising four sequential modules, depicted in Fig. \ref{fig:fig1}, to improve the efficiency of EEG signal generation:

\textbf{Data Preparation}: Following the International 10–20 system, EEG electrode distribution reflects approximately hemispherical symmetry and functional segregation of human brain \cite{hugdahl2005symmetry, wig2017segregated}. Electrodes are initially grouped into distinct electrode regions, then the reference channel signal within each region is selected, and finally is subtracted from other channels within the same region to derive the EEG bias signal for each electrode.

\textbf{Data Preprocessing}: Based on the EEG bias signals derived from Data Preparation module, we commence by empirically eliminating data points within each signal segment deviating by one or two standard deviations from the total segment signal of each channel. Subsequently, linear interpolation is employed to address the missing data points. Finally, Multiscale Principal Components Analysis (MSPCA) \cite{akinduko2014multiscale} is utilized to reduce dimensionality and extract features from the interpolated data. 

\textbf{Biased-EEG Generation}: The Biased-EEG Generation module comprises BiasEEGGanFormer and BiasEEGDiffFormer. BiasEEGGanFormer captures high timing resolution characteristics using Transformer  \cite{vaswani2017attention} with multi-head attention, producing the “One-stage Generated Bias”. In BiasEEGDiffFormer, spatial alignment and feature extraction are first performed by ConvBlock, followed by DiffusionBlock that generates signal samples incrementally from noise data, ensuring consistency with electrode locations, resulting in the “Two-stage Generated Bias”.

\textbf{Synthetic EEG Generation}: In this module, we add the Regional Reference Signal to the Two-stage Generated Bias to generate signals for each channel. Specifically, we adopt divide-and-conquer approach, selecting key channels within each region as references and optimizing the derivation path with greedy algorithm.

In summary, this study contributes significantly in the following two aspects:

\textbf{Formulation of the Dense-channel EEG Generation Problem}: We formulate the problem of generating dense-channel EEG signals from sparse channels for whole brain as a set of cross-channel EEG generation tasks. This formulation addresses challenges related to biased EEG signal generation, EEG electrode position alignment and the synthesis of EEG signals across channels.

\textbf{Development of a Unified Framework for Dense-channel EEG Generation}: We introduce YOAS, a unified framework for dense-channel EEG generation. This framework integrates prior information on electrode positioning from the International 10-20 system, precise bias EEG data for each channel, and general across-channel EEG synthesis paradigms to effectively produce multi-channel EEG signals.

\section{RELATED WORK}
\label{sec:2}
\subsection{Across-channel EEG Signal Generation}
\label{subsec:2.1}
In this study, we delineate the cross-channel EEG signal generation task as the process of generating an additional EEG channel signal based on the collected EEG channel signals. Firstly, previous studies predominantly concentrated on data augmentation, but failed to accomplish cross-channel EEG signal generation. For instance, EEG-GAN \cite{hartmann2018eeg} primarily targets the single-channel data augmentation, showing an enhancement over WGAN \cite{arjovsky2017wasserstein}; Nonetheless, its pursuit of capturing features in lengthy sequences and local intricacies escalates computational costs and training instability significantly. Moreover, although BioDiffusion \cite{li2024biodiffusion} excels in producing high-fidelity multivariate signals, it does not specifically cater to cross-channel EEG signal generation. Consequently, we introduce the Biased-EEG Generation module incorporating Transformer \cite{vaswani2017attention} and diffusion model \cite{sohl2015deep}, \cite{ho2020denoising}, thereby facilitating cross-channel EEG signal generation. 
\subsection{EEG Electrode Position Alignment}
\label{subsec:2.2}
The EEG electrode position alignment refers to the process of aligning generated EEG signal data with the physical positions of the electrodes to which they correspond. Generally speaking, diffusion models \cite{sohl2015deep} operating by incrementally introducing and reducing noise through both forward and reverse processes, is widely used to generate the data. The proposed DDPM \cite{ho2020denoising} further enhances the model, surpassing GAN \cite{dhariwal2021diffusion} in generating high-quality images. While diffusion models have been applied to generate high-fidelity, non-stationary biomedical signals \cite{li2024biodiffusion}, they primarily focus on enhancing single-channel data, not multi-channel or cross-channel data. Thus, a crucial issue that remains unresolved is the lack of adherence to the positional characteristics of the electrodes in the generated signals. In BiasEEGDiffFormer of YOAS, we propose an innovative diffusion model to align electrode positions by matching signals generated by BiasEEGGanFormer with corresponding reference signals in their respectively divided regions.
\subsection{Across-channel EEG Synthetic Paradigm}
\label{subsec:2.3}
The across-channel EEG synthetic paradigm refers to the deduction paradigm process that pursuing the optimized sets of the generation paths when generating multi-channel EEG signals across the whole brain. Most of EEG signal generation studies primarily focused on signal generation for data augmentation \cite{Ajra_2023}, \cite{li2024biodiffusion}. However, in the pursuit of generating multi-channel EEG signals throughout the entire brain, the task of deriving EEG signals for all channels from a single channel signal frequently presents significant challenges. Hence, it becomes imperative to determine and refine the optimal number of reference channel signals required to generate signals across all channels. Additionally, the optimization and delineation of generation pathways for specific channels based on the designated reference channels are also paramount. In response to these challenges, we introduce the Synthetic EEG Generation module within the YOAS framework to effectively address the aforementioned concerns.
\section{METHOD}
\label{sec:3}
\subsection{Overview}
\label{subsec:3.1}
The task for dense-channel EEG signal generation for entire brain using YOAS, as depicted in Fig. \ref{fig:fig1}, could be theoretically formulated by optimizing a set of across-channel EEG signal generation problems based on $K$ reference channels, where each reference channel represents a regional division (RD). Initially, as shown in Data Preparation of Fig. \ref{fig:fig1}, the entire brain positions are partitioned into $K$ RDs according to the widely-used International 10-20 system in EEG, denoted as $RD_i$ for the $i$-th division and $M_i$ as the total number of channels within $RD_i$. Secondly, we can record the channel index within $RD_i$ as set $\hat{M}_{i}=\{1,2,\ldots,j,\ldots, M_i\}$, with $j$ standing for the $j$-th channel ($1<j\leq M_i$) of $RD_i$. Thirdly, the whole brain channel signal set can be formulated as $\{O_{ij}\}_{i=1,j=1}^{K,M_i}$, the reference channel signal set can be defined as $\{O_{i1}\}_{i=1}^{K}$ by selecting the first channel for each RD, and the remaining channel signal set can be represented by $\{O_{ij}\}_{i=1,j=2}^{K,M_i}$. Thus, the across-channel EEG signal generation problem is involved by asking that “Can the remaining channel signals $O_{ij}$ ($1<j\leq M_i$) in the $RD_i$ be accurately generated based on the reference channel $O_{i1}$, subject to $0<i\leq K$?” Mathematically, the dense-channel EEG signal generation problem of whole brain can be initially expressed as follows: 
\begin{equation}
\label{eq1}
Gen_{n}(O_{i1})\to C_{ij}, \\
\end{equation}
\begin{equation}
\label{eq2}
\{Gen\}=\{Gen_{1}, Gen_{2},...,Gen_{n},...,Gen_{N}\},
\end{equation}
where $\{Gen\}$ represents a set of $N$ possible strategy to generate signals, and $C_{ij}$ represents the generated signal of $O_{ij}$, satisfying $i \in [1,K]\;, \;j \in [2,M_i]\;, n \in[1,N]$.

In order to accurately solve Formula \eqref{eq1} and Formula \eqref{eq2}, here we innovatively introduce the variable of the bias signal of each regional division, namely, $RD_{i}$, based on the corresponding reference signal $O_{i1}$ ($0<i\leq K$), which can be calculated as follow:
\begin{equation}
    \label{eq3}
    \{\beta_{ij}\}_{i=1,j=2}^{K,M_i}=\{O_{ij}\}_{i=1,j=2}^{K,M_i}-\{O_{i1}\}_{i=1}^{K},
\end{equation}
where $\beta_{ij}$ represents the bias data of other channel signals in $RD_i$ relative to the reference channel. Subsequently, we could rewrite Formula \eqref{eq1} and Formula \eqref{eq2} respectively as:
\begin{equation}
\label{eq4}
Gen'_{n}(O_{i1},\beta_{ij})\to C_{ij}, \\
\end{equation}
\begin{equation}
\label{eq5}
\{Gen'\}=\{Gen'_{1}, Gen'_{2},...Gen'_{n}...,Gen'_{N}\}.
\end{equation}
To address the optimization of Formulas \eqref{eq4} and \eqref{eq5}, the key issue lies on refining the generation process of bias signals. In this study, the quality of the bias signal entails Data Preprocessing stage as shown in Fig. \ref{fig:fig1} for $\beta_{ij}$, elaborated in Section \ref{subsec:3.3}, resulting in $\beta'_{ij}$. 

Subsequently, as shown in Biased-EEG Generation stage of Fig. \ref{fig:fig1}, we incorporate the temporal attributes of the biased EEG signal and the spatial characteristics of the electrode positions, organizing the generation process into two stages. This entails the utilization of two generation modules, labeled as $Trans$ (illustrated in Fig. \ref{fig:fig3}) and $Diff$ (depicted in Fig. \ref{fig:fig4}), to produce One-stage bias $b_{ij}$ and Two-stage bias $B_{ij}$, respectively. Thus, the procedure is expressed as follows, with details provided in Section \ref{subsec:3.4}:
\begin{align}
\label{eq6}
b_{ij}&=Trans(\beta'_{ij}),\\
\label{eq7}
B_{ij}&=Diff(b_{ij}).
\end{align}

Finally, as displayed in Synthetic EEG Generation stage of Fig. \ref{fig:fig1}, we synthesize the Two-stage bias $B_{ij}$ and the reference channel $O_{1j}$ of the  $RD_{i}$ to generate $C_{ij}$, satisfying $0<i\leq K$ and $0<j\leq M_i$. Taking into account Formula \eqref{eq4}, we further have expressions as follows:
\begin{align}
\label{eq8}
C_{ij}=yoas(B_{ij},O_{i1},Gen'_{n}(O_{i1},B_{ij})),
\end{align}
where the $yoas(.)$ means a final EEG generation operation that $B_{ij}$ is combined with $O_{1j}$ according to the generation path strategy of $Gen'_{n}(O_{i1},B_{ij})$ to generate $C_{ij}$.

Once the channel signal in $RD_i$ has been generated, the corresponding generation path within this division is simultaneously determined. However, the feasibility of merging the corresponding generation paths across regional divisions remains uncertain. Therefore, we execute the merging process between the generated paths and rewrite Formula \eqref{eq8} accordingly as follows:
\begin{equation}
\label{eq9}
X=bool(Gen'_n(O_{m1},\beta_{i1}) \to C_{i1}),
\end{equation}
\begin{equation}
\label{eq10}
C_{ij}=yoas'(B_{ij},\{O_{i1},O_{m1}\},X,Gen'_{n}(O_{i1},B_{ij}),
\end{equation}
\begin{equation}
\begin{aligned}
\label{eq11}
C_{ij}=\hat{yoas}(B_{ij},\{O_{i1},O_{m1}\}\\
Gen'_{n}(O_{i1},B_{ij}), Gen'_{\hat n}(O_{m1},B_{ij})),
\end{aligned}
\end{equation}
\begin{equation}
\label{eq12}
\hat {RD}_t = RD_i\oplus RD_m.   
\end{equation}
where $\hat{K}$ represents the number of merged areas $\hat{RD}_t$, $m_t$ signifies the total number of channels in $\hat{RD}_t$, with $t$ representing the number of areas after merging, and $\oplus$ denotes merge operation. Utilizing Formula \eqref{eq9}, $X$ is obtained through the $bool(.)$ operation, which evaluates whether the reference channel $O_{i1}$ from $RD_i$ can be generated by the reference channel $O_{m1}$ from $RD_m$. If $X=0$, indicating that $RD_m$ and $RD_i$ cannot be merged, Formula \eqref{eq10} reverts to Formula \eqref{eq8}. Conversely, if $X=1$, signifying that $RD_m$ and $RD_i$ can be merged into $\hat{RD}_t$ ($t \leq \hat{K}$) as shown in Formula \eqref{eq12}, then Formula \eqref{eq10} is redefined as Formula $\eqref{eq11}$, where $C_{ij}$ can be generated through $O_{i1}$ and $Gen'_{n}(O_{i1},B_{ij})$, or derived from $O_{m1}$ and $Gen'_{\hat{n}}(O_{m1},B_{ij})$.
\subsection{Data Preparation}
\label{subsec:3.2}
This section elucidates the initial regional division discussed in Section \ref{subsec:3.1}, which divides all brain channels into $K$ RDs based on the International 10–20 system and the approximately structural/functional symmetry between the left and right hemispheres \cite{martinez2019automatic}. To achieve this, we propose three hypotheses for multi-channel EEG generation pathways, illustrated in Fig. \ref{fig:fig2}. For convenience, we denote that $O_{ij}$ represents the source channel, and $O_{im}$ represents the target channel within the $RD_{i}$. Moreover, $Dis( . )$ calculates the physical distance between two electrodes, while $D( . )$ computes the correlation distance between their signals, where $D=\mid1-\rho\mid $ $(-1<\rho<1)$, with $\rho$ representing the Pearson Correlation Coefficient.

\textbf{Hypothesis 1\label{subsubsec:3.2.1}}. In Fig. \ref{fig:fig2}(A), if the criteria\\ $\begin{cases}Dis(O_{ij},O_{im})\leq L_1\\D(O_{ij},C_{im})\leq P_1\end{cases}$ is met, it implies that the signal of the target channel $O_{im}$ can be directly generated through the source channel $O_{ij}$, i.e., $O_{ij}\longrightarrow O_{im}$.

\textbf{Hypothesis 2\label{subsubsec:3.2.2}}. In Fig. \ref{fig:fig2}(B), for a specific channel, if the condition\\ $\begin{cases} \begin{cases}Dis(O_{ij},O_{in})\leq L_2\&Dis(O_{ij},O_{im})\leq L_2\\\&Dis(O_{im},O_{in})\leq L_2\end{cases}\\D(O_{ij},C_{in})\leq P_2\\D(O_{in},C_{im})\leq P_2\end{cases}$ is satisfied, it indicates that the signal of target channel $O_{im}$ can be indirectly generated by that of the source channel ${O_{ij}}$, i.e., $O_{ij}\xrightarrow{o_{in}}O_{im}$ through intermediate channel $O_{in}$. 

\textbf{Hypothesis 3\label{subsubsec:3.2.3}}. In Fig. \ref{fig:fig2}(C), for channels demonstrating strong symmetry between the left and right hemispheres, if the criteria $\begin{cases}Dis(O_{ij},O_{im})\leq L_3\\D(O_{ij},C_{im})\leq P_3\\D(C_{ij},O_{im})\leq P_3\end{cases}$ is satisfied, this signifies a mutual signal generation between the source channel $O_{ij}$ and the target channel $O_{im}$, i.e., $O_{ij}\longleftrightarrow O_{im}$.

To this end, we record the maximum physical radiation generation range set and the coefficient set in the aforementioned three hypotheses as $L=\{L_1,L_2,L_3\}$ and $P=\{P_1,P_2,P_3\}$ $\left.\left(\begin{matrix}0\leq P_i\leq2\\\end{matrix}\right.\right)$, respectively. 
In the International 10-20 system for EEG placement, zoning procedure proceeds as follows: initially, we empirically partition it into 5 regions comprising the Frontal Lobe, Temporal Lobe, Central Sulcus, Parietal Lobe, and Occipital Lobe. Subsequently, leveraging the approximately hemispherical symmetry, we utilize the central sulcus as the axis for division, excluding paired regions with notable signal correlation such as (A1, A2) and (O1, O2). Further, we apply the aforementioned hypotheses to refine the regional division task. Hypothesis 1 (\ref{subsubsec:3.2.1}) sets $L_1$ empirically to the radius of the 2D distribution map with $P_1$ equal to 0.3. Hypothesis 2 (\ref{subsubsec:3.2.2}) maintains $L_2$ at the 2D distribution map's radius, with $P_2$ equal to 0.3. Hypothesis 3 (\ref{subsubsec:3.2.1}) sets $L_3$ to the 2D distribution map's diameter, with $P_3$ equal to 0.1.
\subsection{Data Preprocessing}
\label{subsec:3.3}
Given the low signal-to-noise ratio (SNR) of EEG signals and the potential for signal corruption, it is likely that the bias signal data $\{\beta_{ij}\}_{i=1,j=2}^{K,\hat{M}}$ derived from the Data Preparation stage of Fig. \ref{fig:fig1} may contain erroneous data. To mitigate these challenges, the following preprocessing procedures have been implemented. Initially, outlier removal is conducted by empirically eliminating data points within each signal segment that deviate by one or two standard deviations from the total segment signal of each channel. Subsequently, linear interpolation is applied to address missing data points. Additionally, NaN values in $\{\beta_{ij}\}_{i=1,j=2}^{K,\hat{M}}$ are filled using linear interpolation, and any remaining INF values are replaced with zero. Finally, Multiscale Principal Components Analysis (MSPCA) \cite{akinduko2014multiscale} is employed to reduce dimensionality and extract effective features from the interpolated data.

\subsection{Biased-EEG Generation}
\label{subsec:3.4}
Previous methodologies predominantly focused on signal generation for single-channel data augmentation, thereby exhibiting limitations in the context of dense-channel EEG signal generation. This task presents challenges in terms of efficiency and consideration of electrode position characteristics. To overcome these constraints, we introduce the Biased-EEG Generation module, consisting of two sub-modules: BiasEEGGanFormer (shown in Fig. \ref{fig:fig3}) and BiasEEGDiffFormer (displayed in Fig. \ref{fig:fig4}). 
\begin{figure*}
	\centering
	\includegraphics[width=\linewidth]{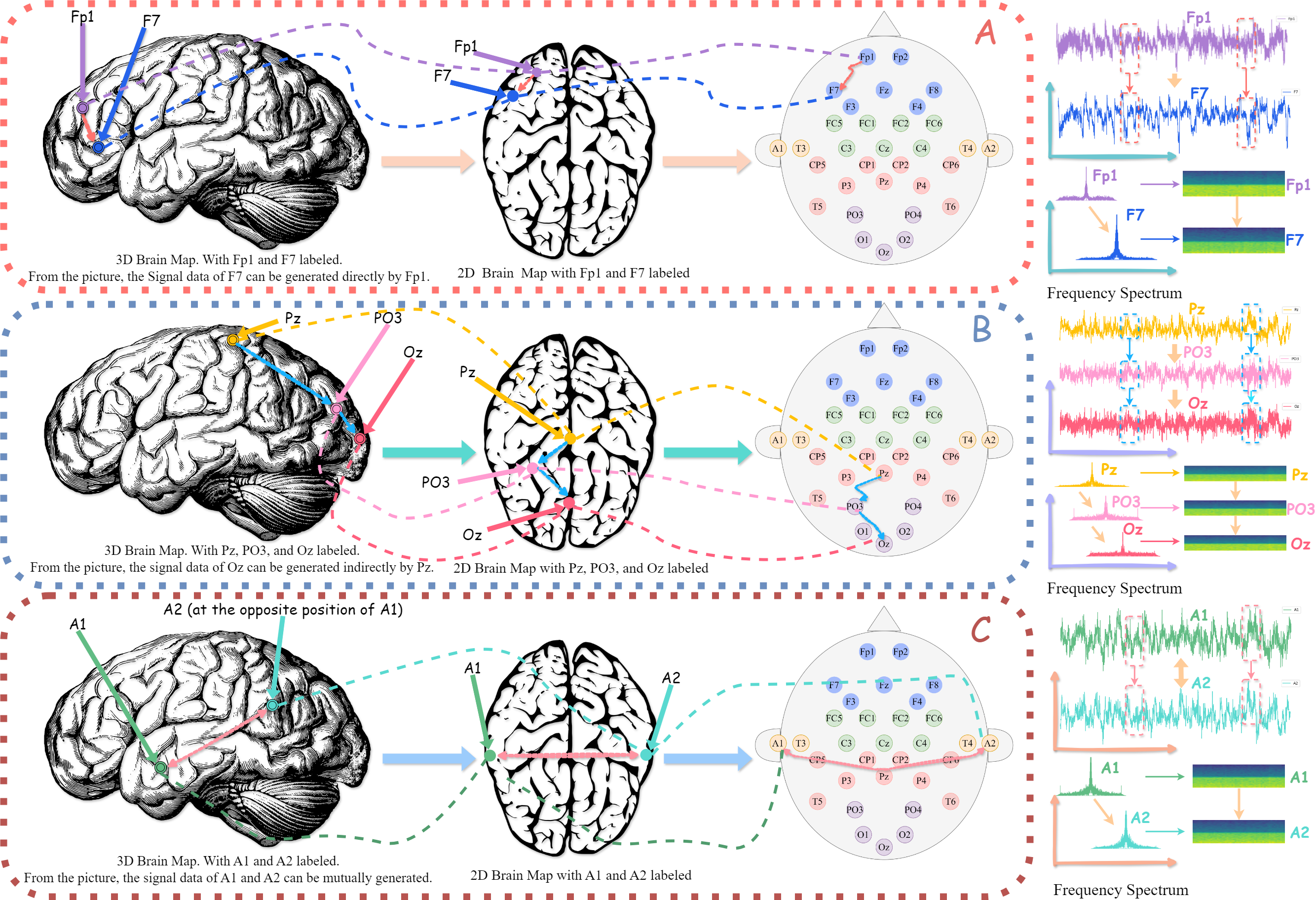}
	\caption{Hypotheses illustration for dense-channel EEG generation pathways: (A) Direct one-way generation involves generating signals directly with reference to the channel; (B) Indirect one-way generation requires an intermediate channel for the reference channel to generate signals indirectly; (C) Mutual generation involves two channels serving as reference channel mutually.}
	\label{fig:fig2}
	\medskip
	\small
\end{figure*}

\begin{figure}
	\centering
	\includegraphics[width=0.95\linewidth]{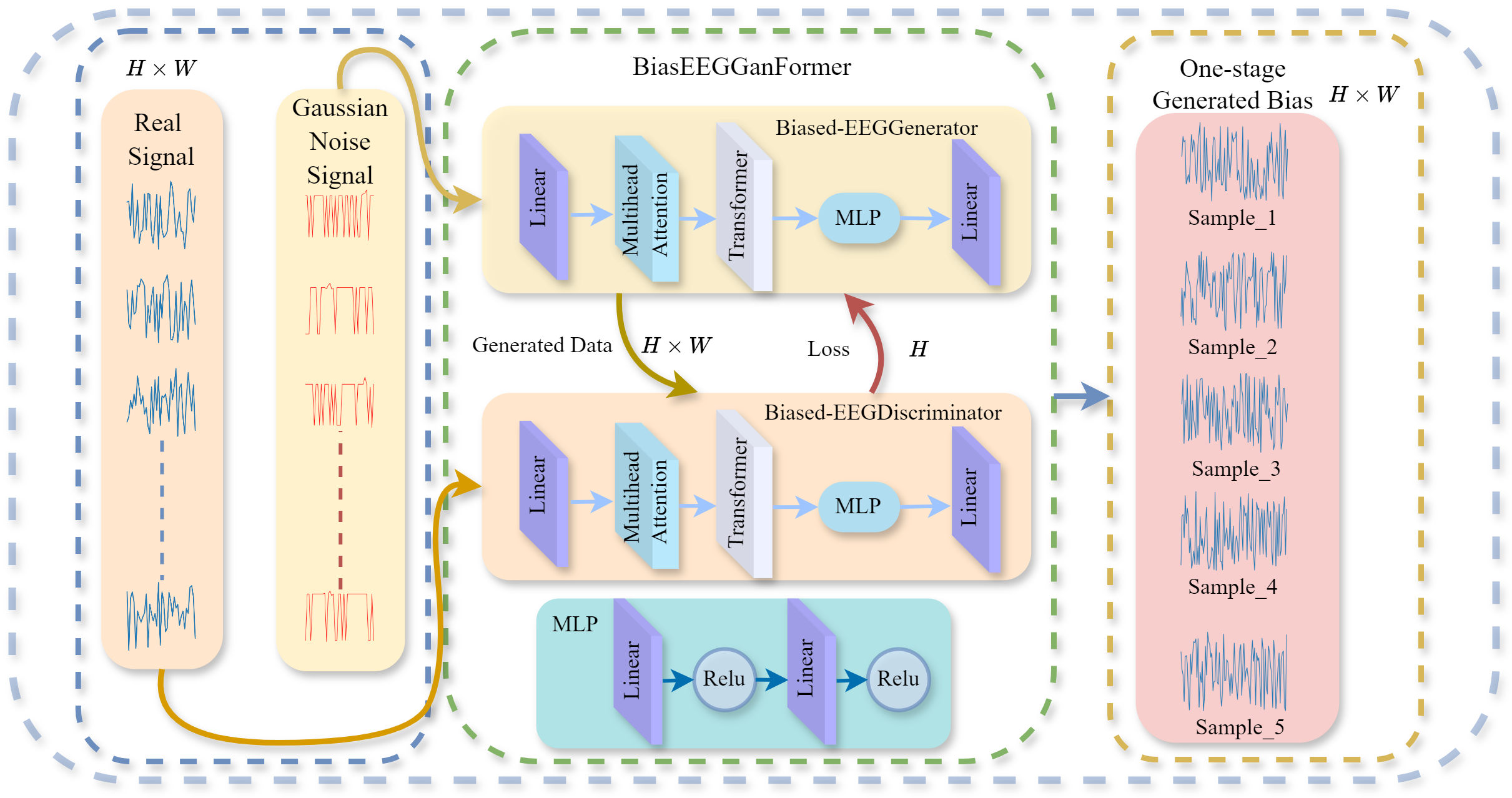}
	\caption{BiasEEGGanFormer employs an adversarial generation architecture, where the Biased-EEGGenerator generates data fed into the Biased-EEGDiscriminator for loss feedback, thereby improving the generation quality. The resulting output comprises the One-stage Generated Bias, enhancing EEG signal refinement.}
	\label{fig:fig3}
	\medskip
	\small %
\end{figure}

\begin{figure}
	\centering
	\includegraphics[width=0.85\linewidth]{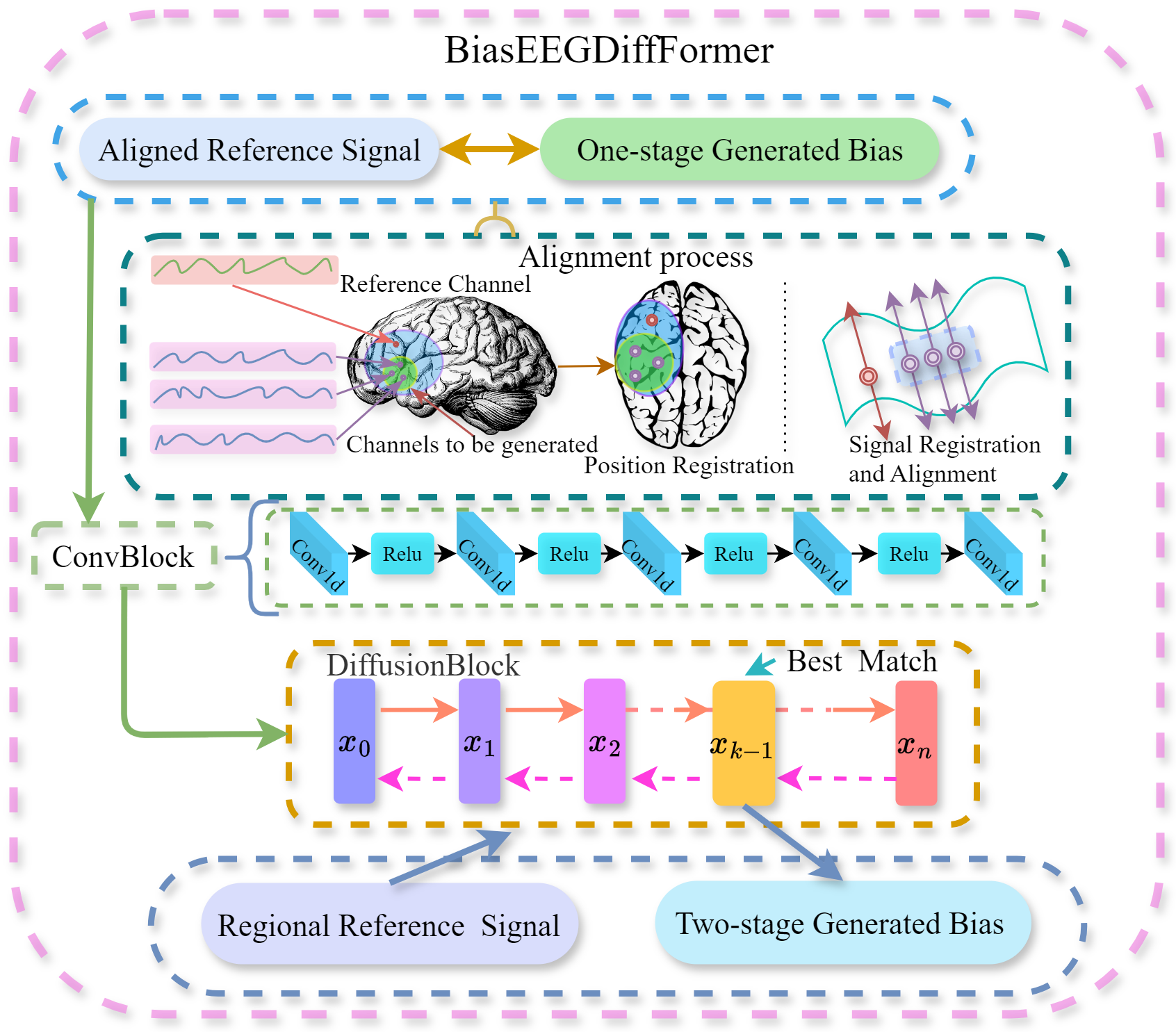}
	\caption{BiasEEGDiffFormer aligns One-stage Generated Bias with the correspondingly physical locations, first applying 1D convolution in ConvBlock for feature extraction, and subsequently involving DiffusionBlock to generate Two-stage Generated Bias based on Regional Reference Signal and diffusion model.}
	\label{fig:fig4}
	\medskip
	\small 
\end{figure}

\begin{figure}
	\centering
	\includegraphics[width=0.85\linewidth]{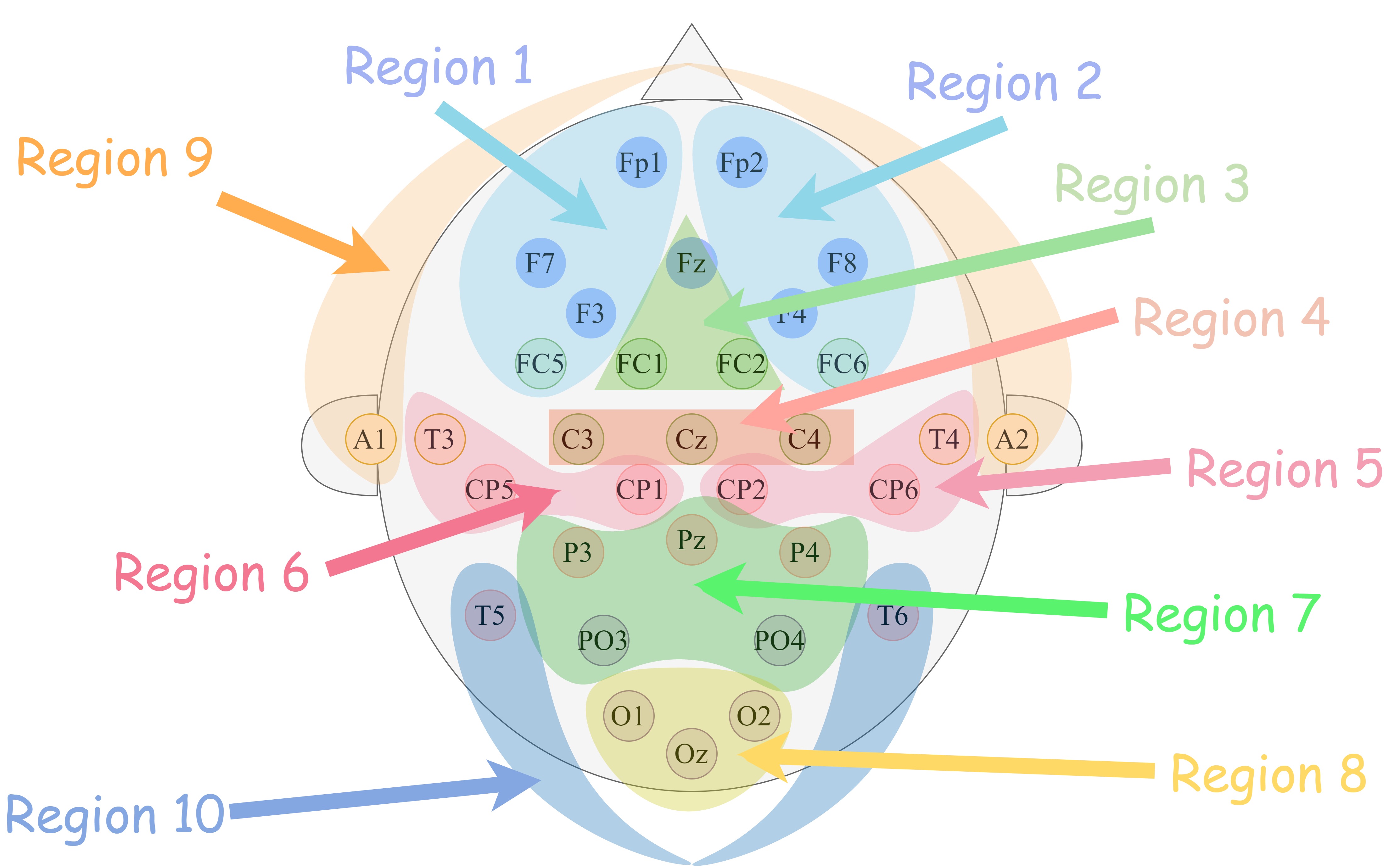}
	\caption{The 32-channel EEG 2D distribution map and the initial EEG channel regional division, with Fp1, Fz, Cz, A1, CP1, CP2, Pz, T5, Oz as reference channels for their respectively regional division.}
	\label{fig:fig5}
	\medskip
	\small
\end{figure}
\textit{{BiasEEGGanFormer.}} Compared with the WGAN series \cite{gulrajani2017improved}, the Transformer model \cite{vaswani2017attention} adopts a self-attention mechanism, enabling it to capture long-term dependencies in time series data without being constrained by sequence length. Compared with traditional GAN models, the Transformer mitigates issues such as generator collapse and training instability \cite{gulrajani2017improved}. To generate quasi-realistic data, we propose BiasEEGGanFormer as shown in Fig. \ref{fig:fig3}, which retains the original GAN architecture while incorporating the Transformer. Specifically, inspired by \cite{zeng2023transformers}, we opt for a simpler approach, adding linear and ReLU layers after the Transformer structure. For the generator, an embedding step extracts features from input data, followed by the Multi-Head Self-Attention module capturing sequence dependencies. Subsequently, an MLP with two linear layers and ReLU activation further processes encoded and decoded outputs, with data generated through a fully connected layer. Both generator and discriminator share the same structure for balance, producing data termed as ‘One-stage Generated Bias’ i.e. $b_{ij}$ described in Formula \ref{eq6}. Unlike previous works \cite{gulrajani2017improved},\cite{schirrmeister2017deep},\cite{dong2018musegan} employing numerous convolutional layers or complex up-sampling and down-sampling techniques, our BiasEEGGanFormer prioritizes the generation effectiveness and efficiency.
\newcommand{\CustomFunction}[2]{\textbf{CustomFunction}($#1$, $#2$)}
\renewcommand{\algorithmicrequire}{\textbf{Input:}}
\renewcommand{\algorithmicensure}{\textbf{Output:}}
\renewcommand{\algorithmicif}{\textbf{if}}
\renewcommand{\algorithmicthen}{\textbf{then}}
\renewcommand{\algorithmicelse}{\textbf{else}}
\renewcommand{\algorithmicelsif}{\textbf{else if}}
\renewcommand{\algorithmicendif}{\textbf{end if}}
\renewcommand{\algorithmicreturn}{\textbf{return}}

\textit{{BiasEEGDiffFormer.}} Taking advantage of the correlation between EEG signals, electrode positions, and reference channel signals, BiasEEGDiffFormer (Fig. \ref{fig:fig4}) is developed as an extension of BiasEEGGanFormer.
Initially, we align each One-stage Generated Bias $b_{ij}$ with the reference signal $O_{i1}$ of its associated physical location $RD_i$. By merging these generated data with reference signals, we construct a virtual two-dimensional EEG distribution map, which enhances the interpretability of generated data $C_{ij}$ for each channel. Next, we utilize the ConvBlock to further extract features, followed by DiffusionBlock \cite{ho2020denoising}. Drawing upon the regional reference signal, we meticulously select the generated data that most accurately aligned with its position, resulting in ‘Two-stage Generated Bias’ i.e. $B_{ij}$ describled in Formula \ref{eq7} with a more detailed expression as
\begin{equation}
\label{eq13}
B_{ij}=Diff(b_{ij},O_{i1},RD_{ij}).
\end{equation}
Typically, the forward process of the diffusion model gradually adds noise to the original signal to generate a noise signal. For each time step $t$, the change in the signal can be expressed as:
\begin{equation}
\label{eq14}
\begin{aligned} B_{i,j,t}&=\sqrt{\bar{\alpha}_{t}}\cdot B_{i,j,0}+\sqrt{1-\bar{\alpha}_{t}}\cdot\epsilon,\\q(B_{t}|B_{i,j,t-1})&=\mathcal{N}(B_{i,j,t};\sqrt{\alpha_{t}}B_{i,j,t-1},(1-\alpha_{t})I),\\B_{i,j,0}&=Diff_{init}(b_{ij},O_{i1},RD_{ij}),\\\bar{\alpha}_{t}&=\prod_{s=1}^{t}\alpha_{s},\\\epsilon&\sim\mathcal{N}(0,I),\end{aligned} 
\end{equation}
where $B_{i,j,0}$ is the initial signal, $\bar{\alpha}_t$ is the cumulative diffusion factor of time step $t$, and $\epsilon$ is the noise of standard normal distribution. Thus, the formulation of the forward diffusion process can be expressed as:
\begin{equation}
\label{eq15}
\begin{aligned}
        q(B_{i,j,t}|B_{i,j,0})&=\mathcal{N}(B_{i,j,t};\sqrt{\bar{\alpha}_t}\cdot Diff_{\mathrm{init}}(b_{ij},O_{i1},RD_i),&\\(1-\bar{\alpha}_t)I).
\end{aligned}
\end{equation}

The backward process is used to gradually recover the original signal from the noisy signal. For each time step $t$, the backward process can be expressed as:
\begin{equation}
\begin{aligned}
\label{eq16}
    p_\theta(B_{i,j,t-1}|B_{i,j,t})=&\mathcal{N}(B_{i,j,t-1};\mu_\theta(B_{i,j,t},t),\sigma_t^2I),\\
    \mu_\theta(B_{i,j,t},B_{i,j,t})=&\frac1{\sqrt{\alpha_t}}\left(B_{i,j,t}-\frac{1-\alpha_t}{\sqrt{1-\bar{\alpha}_t}}\epsilon_\theta(B_{i,j,t},t)\right),
\end{aligned}
\end{equation}
where $\mu_\theta(B_{i,j,t},t)$ is used to predict the mean, and $\sigma_t^2$ is the variance of time step $t$. To train the neural network $\epsilon_\theta$, the mean square error is used as the loss function:
\begin{equation}
\label{eq17}
    L=\mathbb{E}_{t,B_{i,j,0},\epsilon}\left[\|\epsilon-\epsilon_\theta(B_{i,j,t},t)\|^2\right],
\end{equation}
by repeatedly applying the backward denoising process, where the final signal $B_{ij}$ is gradually generated from the noise.
Thus, the generation process of $B_{i,j}$ can be rewritten to Algorithm \ref{alg:G2}.

\begin{algorithm}
\small
\caption{\textbf{: Generation Process of $B_{ij}$} }
\label{alg:G2}
\begin{algorithmic}[h]
\REQUIRE \( b_{ij} \), \( O_{i1} \), \( RD_i \), \(P_i\), \(\hat{t}\), \( T \)
\ENSURE Generated signal \( B_{ij} \)
\STATE \textbf{Initialization:} $B_{i,j,0} = Diff_{\text{init}}(b_{ij}, O_{i1}, RD_i)$\\
\STATE \textbf{Forward Process:} \\ Generate noisy signal \( B_{i,j,t} \) for $t = 1, \ldots, T$:
\[B_{i,j,t} = \sqrt{\bar{\alpha}_t} \cdot B_{i,j,0} + \sqrt{1 - \bar{\alpha}_t} \cdot \epsilon\]
where \( \epsilon \sim \mathcal{N}(0, I) \).
\STATE \textbf{Backward Process:}\\ Iteratively denoise from \( t = T \) to \( t = 1 \):
\begin{align*}
\small
B_{i,j,t-1} &\sim \mathcal{N}(\mu_\theta(B_{i,j,t}, t), \sigma_t^2 I)\\
\mu_\theta(B_{i,j,t}, t) &= \frac{1}{\sqrt{\alpha_t}} \left( B_{i,j,t} - \frac{1 - \alpha_t}{\sqrt{1 - \bar{\alpha}_t}} \epsilon_\theta(B_{i,j,t}, t) \right) \\
\sigma_t^2 &= \frac{1 - \bar{\alpha}_{t-1}}{1 - \bar{\alpha}_t} (1 - \alpha_t) \\
\epsilon_\theta(B_{i,j,t}, t) &\approx \text{NN}(B_{i,j,t}, b_{ij}, O_{i1}, RD_i, t) \\
\textbf{If}\; D_{t}(B_{i,j,t},&O_{i,1}, RD_i)\leq P_{i} \;\&\;\underset{i \in T}{\textbf{argmin}}\;D_i=D_{t}\\\textbf{then} \;\hat{t}&=t
\end{align*}
\textbf{Loss Function:} Compute the loss for training:\\
\quad\quad\quad\quad$L = \mathbb{E}_{t, x_0, \epsilon} \left[ \| \epsilon - \epsilon_\theta(B_{i,j,t}, t) \|^2 \right]$
\STATE \textbf{Return:} \(B_{i,j,\hat{t}} \)
\end{algorithmic}
\end{algorithm}

\begin{algorithm}
\caption{\textbf{: Synthetic Paradigm 1}}
\small
\label{alg:Synthetic_Paradigm_1}
\begin{algorithmic}[h]
\REQUIRE Channel list {{$Ch_1, Ch_2,...,Ch_i,...,Ch_N$}} \par $ N\in Z^{*}\;, i \in Z^{*} \;\&\; 1 \leq i \leq N $ 
\ENSURE Refined generation paths among channels 
\\//Direct,Indirect,Mutual
\STATE \hspace*{-
\algorithmicindent}\textbf{start}
\FOR{$Ch_i$ in Channel list}
\FOR{$Ch_j$ in Channel list $(i\neq j)$ 
            ($j \in Z^{*} \& 1 \leq j \leq N $)} 		
\STATE Set $Ch_i$ as Reference Channel
\IF{$Ch_i\to Ch_j\& Ch_i\to Ch_{j+n}$ 
\par \& $ n\in([1-j,-1]\cup[1,N-j]), n \in Z$}
\STATE $Ch_i\to Ch_j\mid Ch_{j+n}$ \; //Hypothesis 1(\ref{subsubsec:3.2.1})
\ELSIF{$Ch_i\to Ch_j\& Ch_j\to Ch_m\;\;(i\neq j\neq m)$}
\STATE $Ch_i\to Ch_m$ \quad\quad\;\; //Hypothesis 2(\ref{subsubsec:3.2.2})
\ELSIF{$Ch_i\to Ch_j\& Ch_j\to Ch_i$}
\STATE $Ch_i\leftrightarrow Ch_j$ \;\;\;\;\;\quad //Hypothesis 3(\ref{subsubsec:3.2.3})
\ELSE
\RETURN "Not feasible" \;\;\;
\ENDIF
\ENDFOR
\ENDFOR
\\
\end{algorithmic}
\end{algorithm}

\begin{algorithm}
\caption{\textbf{: Synthetic Paradigm 2}}
\label{alg:Synthetic_Paradigm_2} 
\begin{algorithmic}[]
\small
\REQUIRE Channel list {{$Ch_1, Ch_2,...,Ch_i,...Ch_N$}
\\ $ N\in Z^{*}\; i \in Z^{*} \;\&\; 1 \leq i \leq N $ } 
\ENSURE Merge results 
\STATE \textbf{start}
\IF{Reference Channel of Regional Division A is $Ch_i$ \\ \& Reference Channel of Regional Division B is $Ch_j$ 
\\ \& $Ch_i\to Ch_j \;\& \; Ch_j\to Ch_i$}
\STATE Merge Regional Division A and Regional Division B
\ELSE 
\RETURN "No Merge"
\ENDIF
\\
\end{algorithmic}
\end{algorithm}

\subsection{Synthetic EEG Generation}
\label{subsec:3.5}
As shown in Fig. \ref{fig:fig1}, to pursue optimized generation paths for generating dense-channel EEG signals across the whole brain, we propose a deduction paradigm consisting of two consecutive paradigms: i.e, \textbf{Synthetic Paradigm 1} (referring to Algorithm \ref{alg:Synthetic_Paradigm_1}) and \textbf{Synthetic Paradigm 2} (referring to Algorithm \ref{alg:Synthetic_Paradigm_2}). Given the initial regional division's insufficient accuracy in Section \ref{subsec:3.2} Data Preparation, \textbf{Synthetic Paradigm 1} aims to refine signal generation paths, including direct, indirect, and mutual generation among channels. Building upon the refined results of \textbf{Synthetic Paradigm 1}, \textbf{Synthetic Paradigm 2} is designed to retrieve the final optimized sets of the generation paths and regional divisions for generating dense-channel EEG signals across the whole brain.

\section{EXPERIMENTS}
\label{sec:4}
\subsection{Implementation Details}
\label{subsec:4.1}
\textbf{Datasets}: We conduct training and validation of our models on the FACED dataset provided in reference \cite{chen2023large}. The dataset comprises recordings from 123 subjects, each with 32 EEG channels, while they watched 28 video clips, all lasting  for 30 seconds. These clips cover nine emotions, including Anger, Disgust, Fear, Sadness, Neutral, Amusement, Inspiration, Joy and Tenderness. Each positive and negative emotion is represented by three clips, while the neutral emotion is depicted in four clips. In the experiments, we select 250Hz as the sampling frequency, resulting in a total of 55 subjects, including 30 subjects for training, and 25 subjects for testing. 

\textbf{Training Details}: We conduct training and testing on the provided dataset using the same parameters outlined in Table \ref{tab:table1}. Initially, for the data utilized during training, we employ SpecGAN \cite{Donahue2018AdversarialAS}, WaveCycleGAN-GP \cite{8639636} and WGAN-GP\cite{gulrajani2017improved} respectively to assess the feasibility of each generation path showed in Fig. \ref{fig:fig5} through the testing of one-to-one channel signal generation. Subsequently, we retest the performance of the paths generated by our BiasEEGGanFormer. All computations are executed on NVIDIA Tesla V100, and the Early-Stopping strategy is applied in the training stage.


\begin{table}[h]
\centering
\caption{Parameters setting for the training/testing stage of YOAS.}
\begin{tabular}{cc}
\hline
\textbf{Parameter} & \textbf{Value} \\ \hline
Input size & 7500 \\ 
Output size & 7500 \\ 
Hidden size & 512 \\ 
Optimizer & Adam \\ 
Learning rate & 0.0001 \\ 
Number of epochs with early stopping & 10000 \\ 
Batch size & 1024 \\ 
Learning rate decay & 0.95 \\ 
Layer Number & 6 \\ 
Head Number & 8 \\ \hline
\label{tab:table1}
\end{tabular}
\end{table}

\subsection{Performance Comparison}
\label{subsec:4.2}
\begin{figure}
\centering
\includegraphics[width=0.98\linewidth]{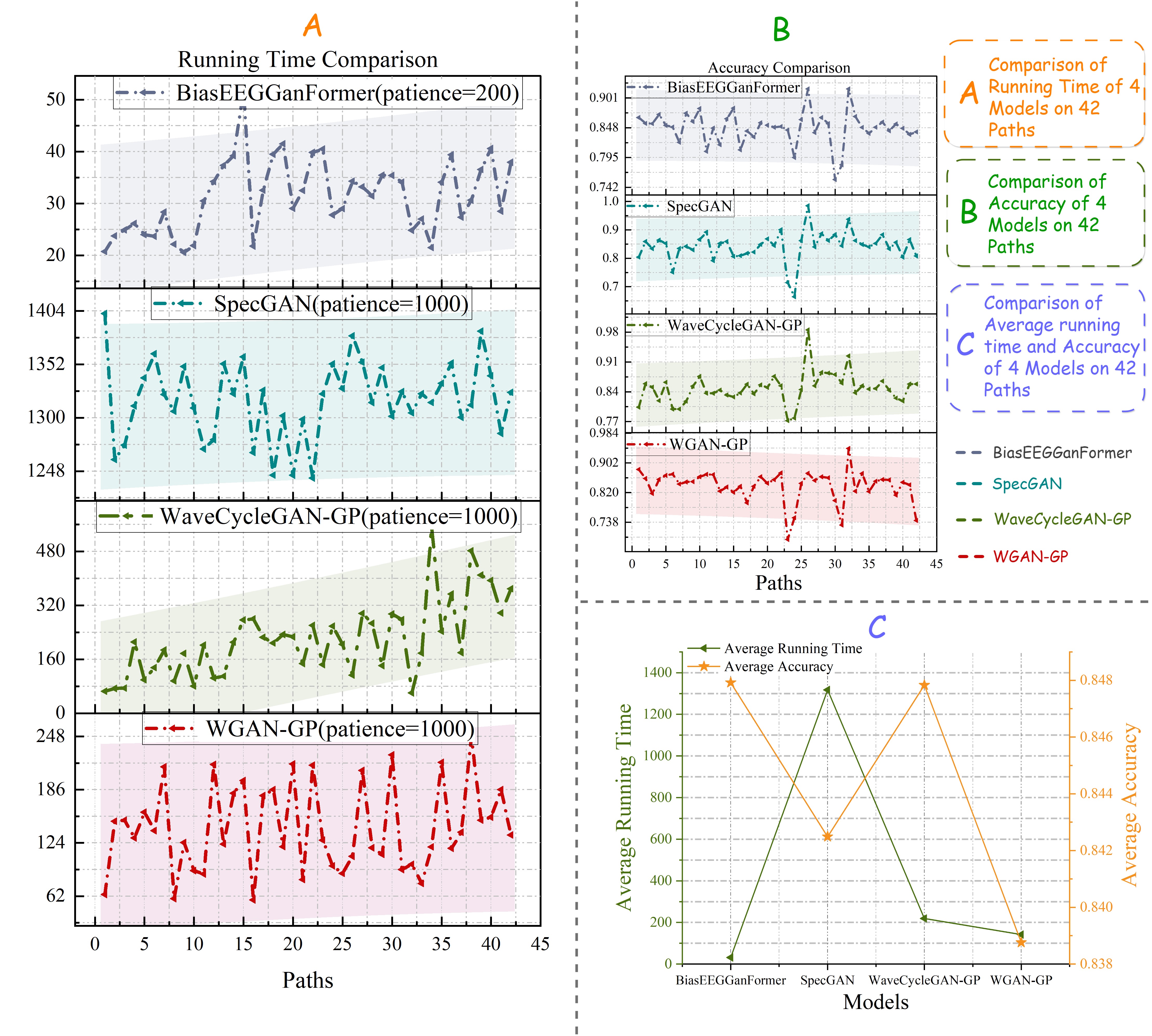}
\caption{Performance comparison between BiasEEGGanFormer, SpecGAN \cite{Donahue2018AdversarialAS}, WaveCycleGAN-GP \cite{8639636} and WGAN-GP\cite{gulrajani2017improved}: (A) and (B) demonstrate that BiasEEGGanFormer is swifter to train than other models while maintaining comparable generation quality. (C) provides further evidence, showing that BiasEEGGanFormer outperforms other models in quality across 42 generation path experiments, with less time cost.}
\label{fig:fig6}
\end{figure}

\begin{figure}
\centering
\includegraphics[width=\linewidth]{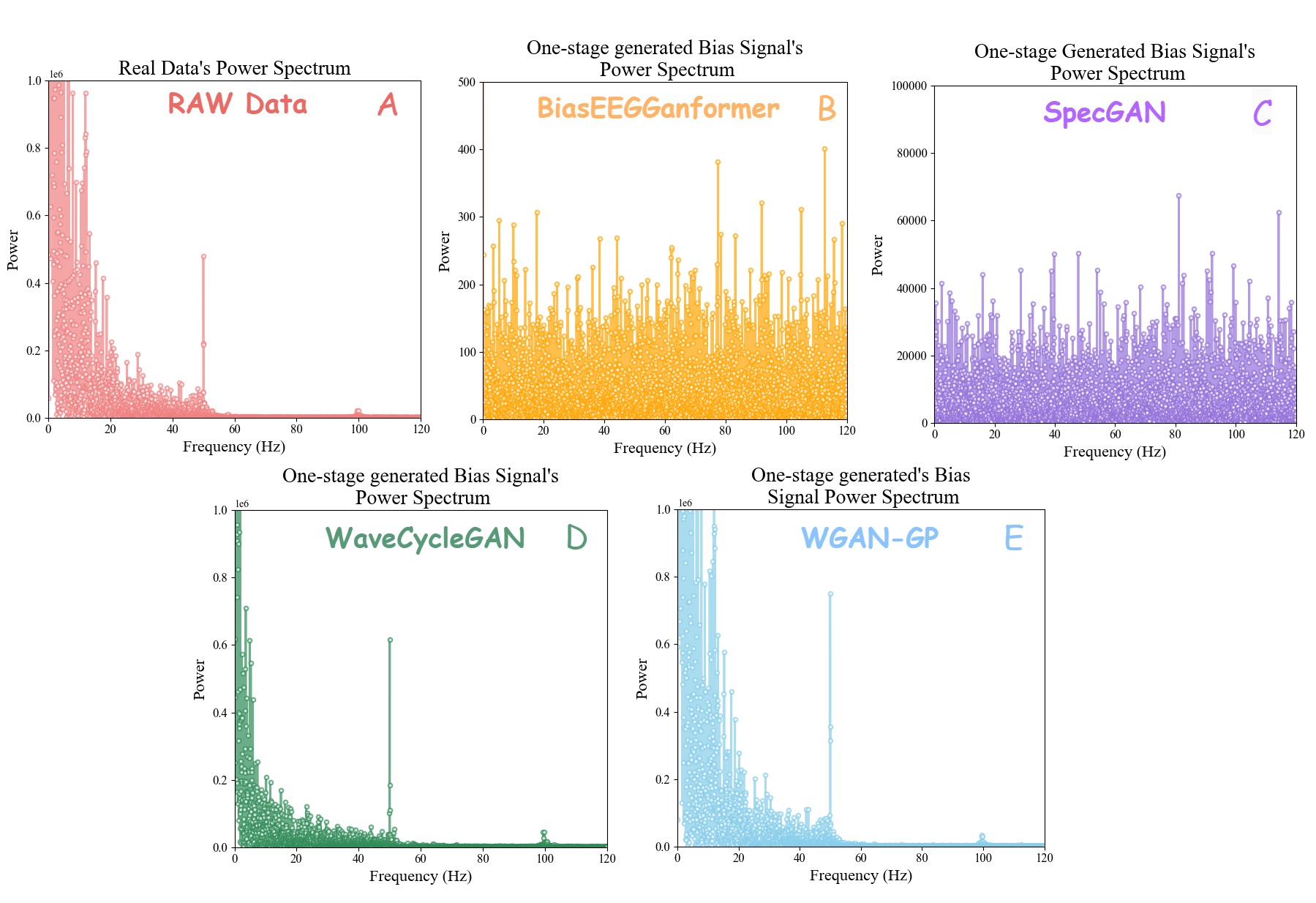}
\caption{The power spectrum comparison between the “indifferent generated signal” of BiasEEGGanformer in (B), SpecGAN \cite{Donahue2018AdversarialAS} in (C), WaveCycleGAN-GP\cite{8639636} in (D), and WGAN-GP\cite{gulrajani2017improved} in (E), demonstrates BiasEEGGanformer's superior performance in attenuating high-amplitude components in frequency-spectrum domain, potentially indicative of noise.}
\label{fig:fig7}
\end{figure}

Based on the provided generation paradigms, we first obtained 10 initial regional divisions for the 32 channels of the International 10–20 system EEG placement, as illustrated in Fig. \ref{fig:fig5}. Further, Fig. \ref{fig:fig6} depicts a comparison of convergence speed between our BiasEEGGanFormer, SpecGAN\cite{Donahue2018AdversarialAS}, WaveCycleGAN\cite{8639636} and WGAN-GP\cite{gulrajani2017improved} across 42 paths, both with Early-Stopping. Notably, BiasEEGGanFormer's patience parameter is set to 200, while SpecGAN, WaveCycleGAN, WGAN-GP's patience is 1000 respectively, and specifically gradient penalty(GP) is applied to WaveCycleGAN. In Fig. \ref{fig:fig6}, BiasEEGGanFormer's runtime is merely 22.22$\%$ of WGAN-GP's, and much shorter than other model's, while maintaining comparable average accuracy. In certain instances, BiasEEGGanFormer even surpasses other models in signal generation quality. Moreover, Fig. \ref{fig:fig7} offers deeper insights into the generation path from O1 to O2, serving as a compelling example highlighting BiasEEGGanFormer's effectiveness in eliminating high-amplitude components in frequency-spectrum domain, thereby enhancing signal quality. In contrast, models compared primarily reproduce the original signal without improvements.

\subsection{Optimized Generation Paths on Neutral Emotion Data}
\label{subsec:4.3}
\begin{figure}
\centering
\includegraphics[width=0.9\linewidth]{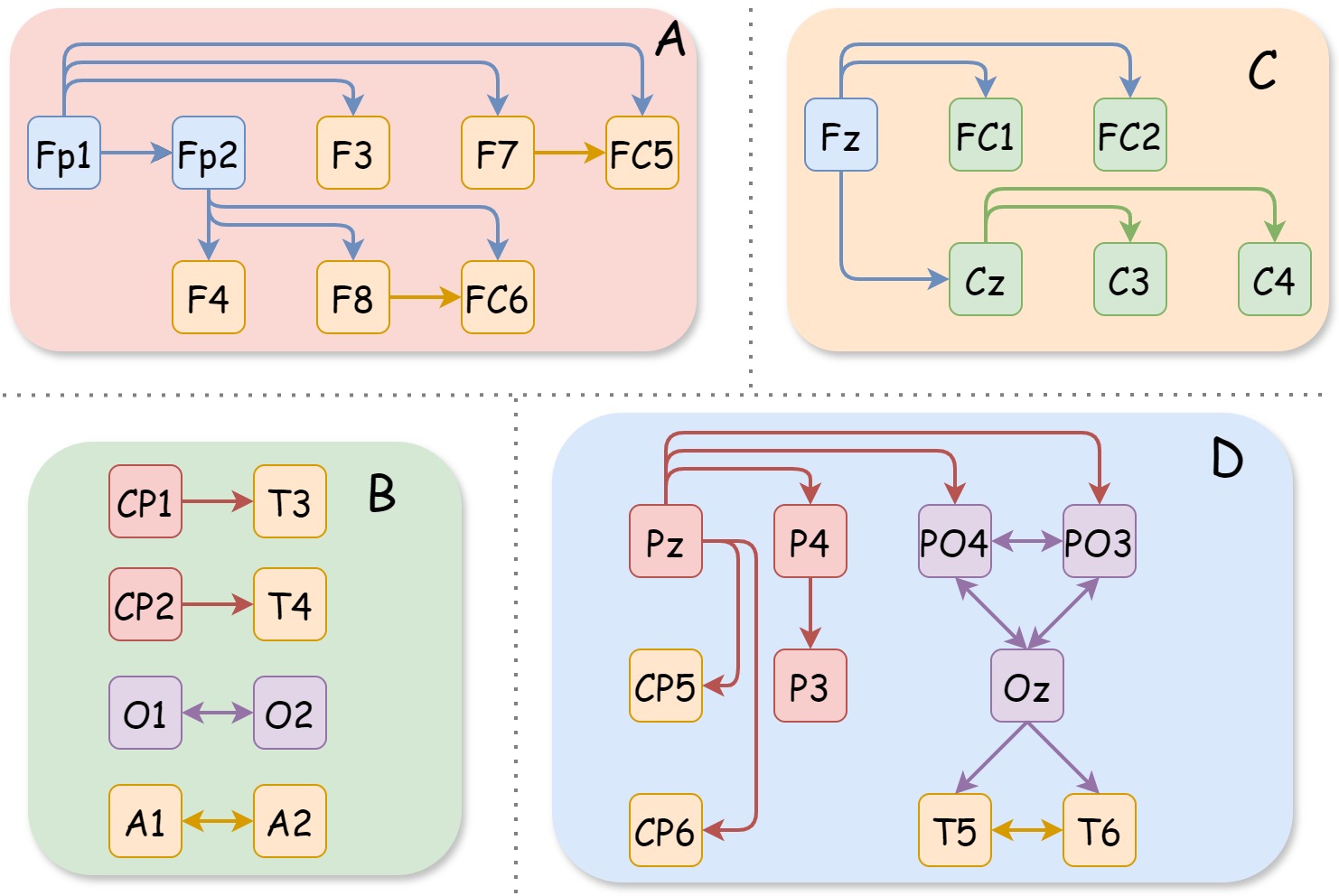}
\caption{Multi-channel EEG signal generation paradigm on neutral emotion data. (A), (C), and (D) are the generation paradigms with Fp1, Fz, and Pz as reference channels, respectively. (B) shows that CP1 and CP2 as reference channels in one-way generation paradigms alone, while pair of O1 and O2 and pair of A1 and A2 are the mutual generation paradigms.}
\label{fig:fig8}
\end{figure}

Based on the output of the YOAS framework in our experiments, we observed that the initial ten regional divisions depicted in Fig. \ref{fig:fig5} were optimized and merged into seven regional divisions on neutral emotion data, as demonstrated in Fig. \ref{fig:fig8} following \textbf{Synthetic Paradigm 2}. Specifically, as illustrated in Fig. \ref{fig:fig8}(A), Fp1 can directly generate F3, F7, FC5, and Fp2, while FP2 indirectly generates F4, F8, and FC6 through Fp1. Notably, F7 can directly generate FC5, and F8 can directly generate FC6. In Fig. \ref{fig:fig8}(B), CP1 generates T3 exclusively, CP2 generates T4 exclusively, while pair of O1 and O2 reciprocally generate each other, as well as pair of A1 and A2, can generate each other mutually. Referring to Fig. \ref{fig:fig8}(C), Fz directly generates FC1, FC2, and Cz, while indirectly generating C3 and C4 through Cz. As shown in Fig. \ref{fig:fig8}(D), Pz directly generates Cp5, Cp6, P4, and PO4, with indirect generation of P3 through P4. Notably, PO4, PO3, and Oz mutually generate each other, with Oz also generating T5 and T6, both of which mutually generate each other.

\begin{figure*}
\centering
\includegraphics[width=\linewidth]{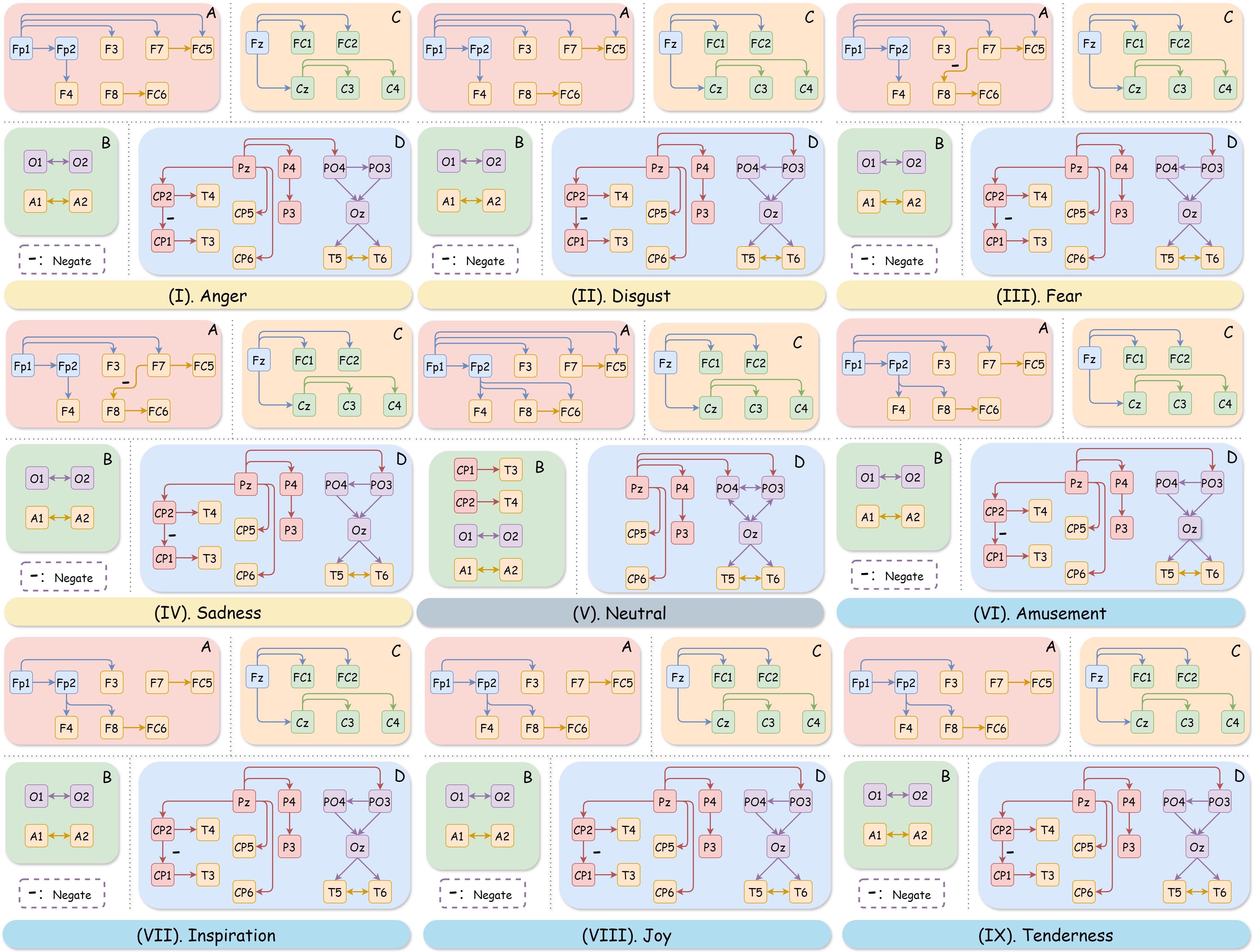}
\caption{Multi-channel EEG signal generation paradigms on neutral, negative and positive emotion data. (I)-(IV) are the optimized generation paths for negative emotion, depicting generation paths of anger, disgust, fear, and sadness respectively. (V) is the optimized generation paths for neutral emotion. (VI)-(IX) are the optimized generation paths for positive emotion, portraying generation paths of amusement, inspiration, joy, and tenderness, in that order.}
\label{fig:fig9}
\end{figure*}

\subsection{Optimized Generation Paths on Negative Emotion Data}
\label{subsec:4.4}
In Fig. \ref{fig:fig9}, (I)--(IV) are the negative emotion generation paradigms. It is worth noting that (I) and (II) share similar generation paradigms, as do (III) and (IV). In (A) of (I)-(IV) is mainly with Fp1 as reference channel, and F8 as reference channel in one-way generation paradigm alone, while in (A) of (III) and (IV), F8 can be derived by negating the F7 signal value generated from Fp1, which further reduces the number of reference channels. In (I)-(IV), CP1 and CP2 originally belonging to (B) in Fig. \ref{fig:fig8} are merged into (D), while (C) remains unchanged. In (D), CP2 can be generated by Pz, and the signal data of CP2 can be inverted to obtain CP1.
\subsection{Optimized Generation Paths on Positive Emotion Data}
\label{subsec:4.5}
In Fig. \ref{fig:fig9}, (VI)--(IX) are the positive emotion generation paradigms. It's also noteworthy that (VII)-(IX) exhibit similar generation paradigms. In (A) of (VI)-(IX), Fp2 serves predominantly as the reference channel, while F7 operates as the reference channel in a one-way generation paradigm alone, contrasting with the structure of (A) in section \ref{subsec:4.4}, but the paradigms in other sections align consistently with section \ref{subsec:4.4}.
\begin{table*}[h]
\centering
\caption{The results of competing models for categorizing the across-subject and intra-subject nine emotions on the original dataset as well as the generated data, Where Raw\_data (\textbf{RAD}) denotes the original dataset, and Generated\_division (\textbf{GD}) denotes data generated based on the generative paths (depicted in Fig. \ref{fig:fig8} and \ref{fig:fig9}) of neutral as well as both negative and positive sentiment generation paths. CL denotes the classification task with cross or intra subject level.}
\begin{tabular}{c|c|c|cccc}
\hline
\textbf{Model} & \textbf{CL} & \textbf{Data} & \textbf{ACC}          & \textbf{AUC}  & \textbf{SEN}          & \textbf{SPE}          \\ \hline
\multirow{4}*{SVM\cite{hearst1998support} } & \multirow{2}*{Cross}       & \textbf{RAD}    & 0.430$\pm$0.051  & 0.675$\pm$0.029  & 0.421$\pm$0.051& 0.424$\pm$0.050    \\ 
               &  & \textbf{GD}& \textbf{0.843$\pm$0.028} & \textbf{0.886$\pm$0.087} & \textbf{0.841$\pm$0.026} & \textbf{0.845$\pm$0.040} \\ \cline{3-7} 
               & \multirow{2}*{Intra} & \textbf{RAD}  & 0.700$\pm$0.030   & 0.828$\pm$0.017          & 0.695$\pm$0.031          & 0.698$\pm$0.031          \\  
               &   & \textbf{GD}  & \textbf{0.940$\pm$0.052} & \textbf{0.966$\pm$0.029} & \textbf{0.939$\pm$0.051} & \textbf{0.940$\pm$0.050} \\ \hline
\multirow{4}*{RandomForest\cite{breiman2001random} }  & \multirow{2}*{Cross}  & \textbf{RAD} & 0.381$\pm$0.049   & 0.649$\pm$0.028  & 0.375$\pm$0.050 & 0.370$\pm$0.057\\ 
               &   & \textbf{GD}  & \textbf{0.947$\pm$0.038} & \textbf{0.970$\pm$0.020} & \textbf{0.947$\pm$0.035} & \textbf{0.948$\pm$0.033} \\ \cline{3-7} 
               & \multirow{2}*{Intra}& \textbf{RAD}  & 0.968$\pm$0.061  & 0.982$\pm$0.035   & 0.968$\pm$0.062          & 0.968$\pm$0.060          \\ 
               &  & \textbf{GD}   & \textbf{0.973$\pm$0.056} & \textbf{0.985$\pm$0.032} & \textbf{0.972$\pm$0.056} & \textbf{0.973$\pm$0.055} \\ \hline
\multirow{4}*{Adaboost\cite{schapire2013explaining} }  & \multirow{2}*{Cross}  & \textbf{RAD} & 0.346$\pm$0.058 & 0.631$\pm$0.032 & 0.343$\pm$0.056 & 0.331$\pm$0.059 \\ 
               &  & \textbf{GD} & \textbf{0.450$\pm$0.021} & \textbf{0.631$\pm$0.031} & \textbf{0.432$\pm$0.018}  & \textbf{0.354$\pm$0.059} \\ \cline{3-7} 
               & \multirow{2}*{Intra} & \textbf{RAD} & \textbf{0.473$\pm$0.018} & \textbf{0.702$\pm$0.010}    & \textbf{0.470$\pm$0.018}     & \textbf{0.463$\pm$0.018}     \\ 
               &  & \textbf{GD}& 0.460$\pm$0.023 & 0.624$\pm$0.037   & 0.441$\pm$0.024  & 0.396$\pm$0.114   \\ \hline
\multirow{4}*{Xgboost\cite{Chen:2016:XST:2939672.2939785} }  & \multirow{2}*{Cross} & \textbf{RAD}  & 0.374$\pm$0.046 &0.646$\pm$0.026 & 0.371$\pm$0.047 & 0.363$\pm$0.052 \\ 
               & & \textbf{GD}& \textbf{0.913$\pm$0.022} & \textbf{0.950$\pm$0.013} & \textbf{0.911$\pm$0.023} & \textbf{0.914$\pm$0.021} \\ \cline{3-7} 
               & \multirow{2}*{Intra}       & \textbf{RAD}           & 0.806$\pm$0.042          & 0.890$\pm$0.024         & 0.804$\pm$0.043 & 0.807$\pm$0.042  \\  
               && \textbf{GD}  & \textbf{0.944$\pm$0.067} & \textbf{0.968$\pm$0.038} & \textbf{0.943$\pm$0.068} & \textbf{0.946$\pm$0.065} \\ \hline
\multirow{4}*{Naive Bayes}  & \multirow{2}*{Cross}       & \textbf{RAD}           & 0.313$\pm$0.033 & 0.613$\pm$0.019  & 0.311$\pm$0.034 & 0.358$\pm$0.053          \\ 
               & & \textbf{GD}& \textbf{0.906$\pm$0.027} & \textbf{0.947$\pm$0.015} & \textbf{0.906$\pm$0.027} & \textbf{0.910$\pm$0.023} \\  \cline{3-7} 
               & \multirow{2}*{Intra}       & \textbf{RAD}      & 0.364$\pm$0.014         & 0.642$\pm$0.008          & 0.363$\pm$0.014  & 0.438$\pm$0.025          \\ 
               & & \textbf{GD}  & \textbf{0.911$\pm$0.050} & \textbf{0.950$\pm$0.028} & \textbf{0.911$\pm$0.049} & \textbf{0.914$\pm$0.048} \\ \hline
\multirow{4}*{ResNet\cite{he2016deep}}   & \multirow{2}*{Cross} & \textbf{RAD}  & 0.577$\pm$0.167     & 0.890$\pm$0.104   & 0.576$\pm$0.166  & 0.614$\pm$0.183  \\ 
               & & \textbf{GD}  & \textbf{0.951$\pm$0.053} & \textbf{0.996$\pm$0.006} & \textbf{0.950$\pm$0.056} & \textbf{0.963$\pm$0.040} \\ \cline{3-7} 
               & \multirow{2}*{Intra}   & \textbf{RAD}           & 0.844$\pm$0.223         & 0.953$\pm$0.089          & 0.843$\pm$0.222   & 0.850$\pm$0.212       \\ 
               && \textbf{GD}  & \textbf{0.965$\pm$0.015} & \textbf{0.998$\pm$0.003} & \textbf{0.963$\pm$0.016} & \textbf{0.967$\pm$0.017} \\ \hline
\multirow{4}*{EfficientNet\cite{tan2019efficientnet}}   & \multirow{2}*{Cross}  & \textbf{RAD}& 0.601$\pm$0.111 & 0.923$\pm$0.071 & 0.607$\pm$0.112  &0.664$\pm$0.133 \\  
               & & \textbf{GD}& \textbf{0.953$\pm$0.069} & \textbf{0.993$\pm$0.013} & \textbf{0.952$\pm$0.070} & \textbf{0.963$\pm$0.054} \\ \cline{3-7} 
               & \multirow{2}*{Intra}   & \textbf{RAD}           & 0.812$\pm$0.183  & 0.960$\pm$0.072  & 0.818$\pm$0.112          & 0.842$\pm$0.153  \\  
               &  & \textbf{GD} & \textbf{0.966$\pm$0.013} & \textbf{0.994$\pm$0.006} & \textbf{0.964$\pm$0.013} & \textbf{0.967$\pm$0.015}  \\ \hline
\end{tabular}
\label{tab:table2}
\end{table*}

\begin{figure}
\centering
\includegraphics[width=\linewidth]{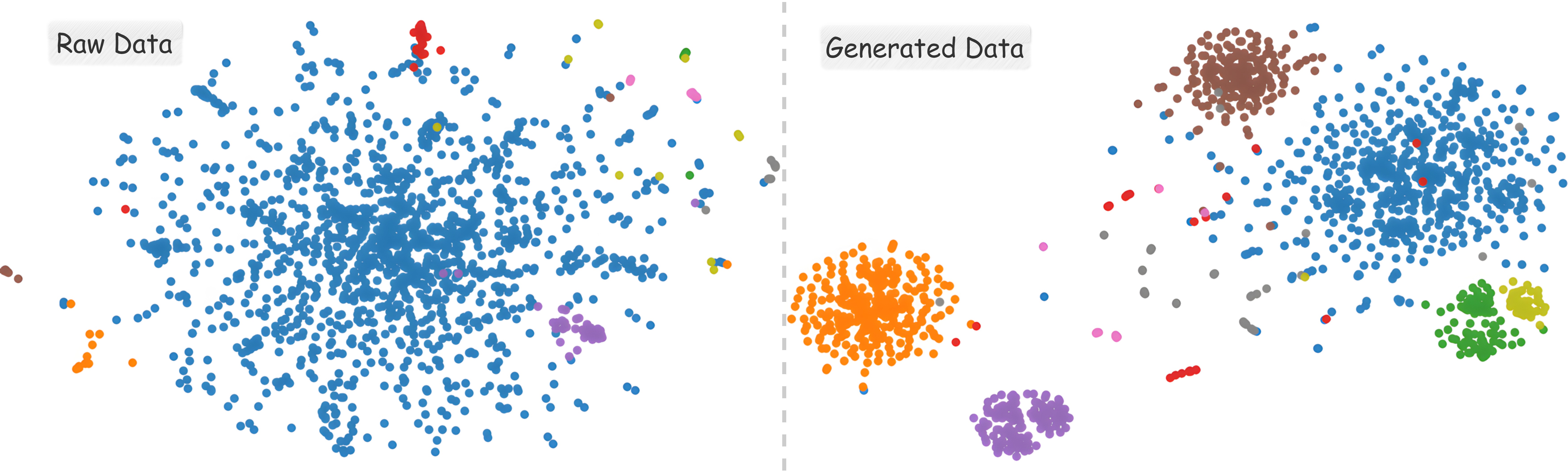}
\caption{Visualization of spectral clustering results using t-SNE based on the raw data \textbf{RAD} and generated data \textbf{GD}.}
\label{fig:fig10}
\end{figure}

\begin{table}[h]
\centering
\caption{The list of the combinations No.2-17 is presented in the form of "A-B-C", where A represents a negative emotion, B represents the neutral emotion, and C represents a positive emotion. }
\begin{tabular}{c|c|c|c}
\hline
\textbf{NO.} & \textbf{Combination} & \textbf{NO.} & \textbf{Combination}    \\ \hline
1 &   Raw\_Data (\textbf{RAD})  &10& Fear-Neutral-Amusement \\
2 &   Anger-Neutral-Amusement  &11& Fear-Neutral-Inspiration \\
3 &   Anger-Neutral-Inspiration  &12& Fear-Neutral-Joy \\
4 &   Anger-Neutral-Joy   &13& Fear-Neutral-Tenderness\\ 
5 &   Anger-Neutral-Tenderness &14& Sadness-Neutral-Amusement\\
6 &   Disgust-Neutral-Amusement  &15& Sadness-Neutral-Inspiration\\
7 &   Disgust-Neutral-Inspiration  &16& Sadness-Neutral-Joy\\
8 &   Disgust-Neutral-Joy  &17& Sadness-Neutral-Tenderness \\
9 &   Disgust-Neutral-Tenderness  &18& Generated\_division (\textbf{GD})\\
\hline
\end{tabular}
\label{tab:table3}
\end{table}

\begin{figure}
\centering
\includegraphics[width=0.98\linewidth]{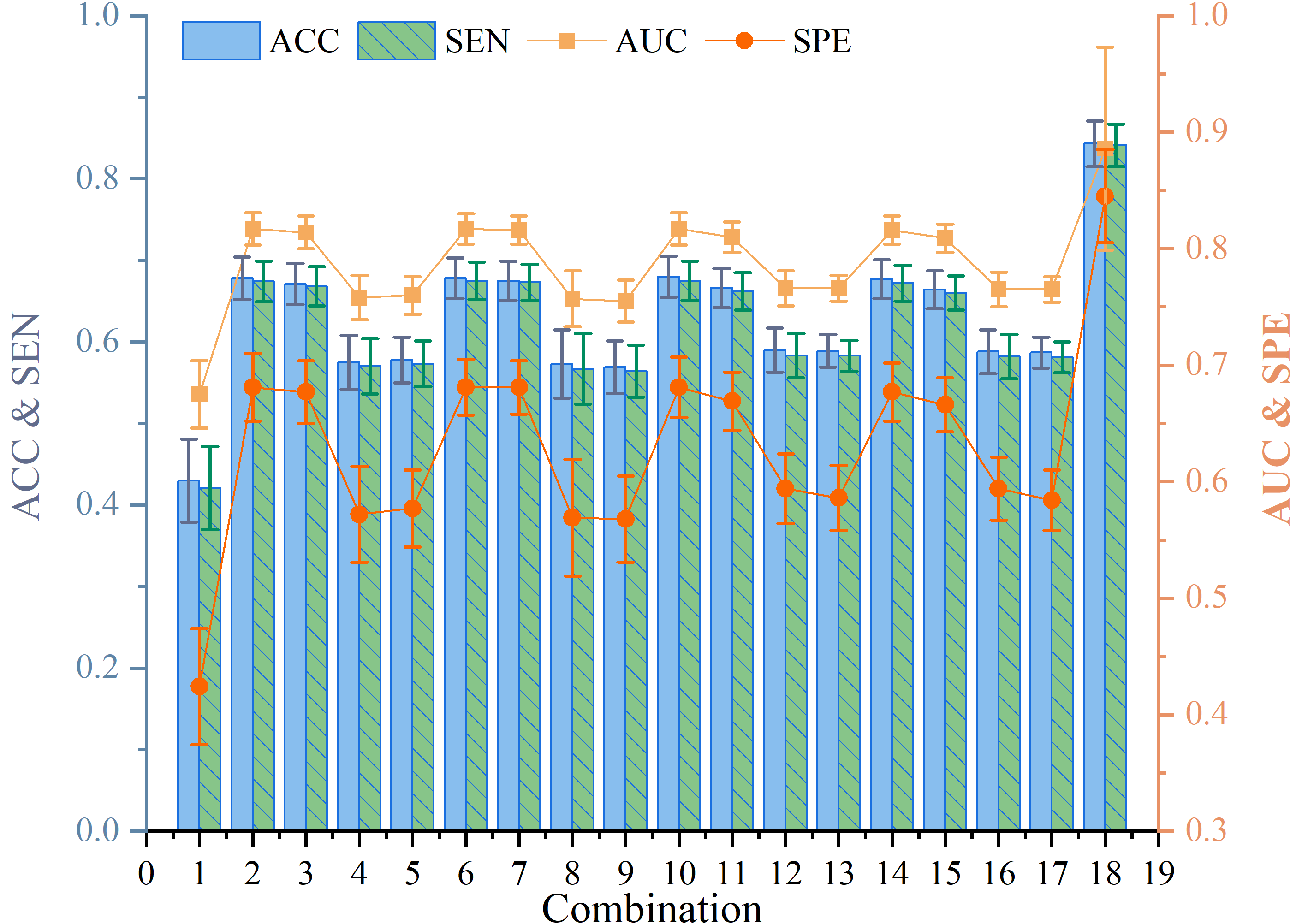}
\caption{The classification results using SVM with DE features to perform cross-subject recognition task of nine-category emotions on combination (1) \textbf{RAD}, combination (2)-(17) and combination (18) \textbf{GD} corresponding to Table \ref{tab:table3} No.1-18 respectively.}
\label{fig:fig11}
\end{figure}

\begin{figure}
\centering
\includegraphics[width=0.95\linewidth]{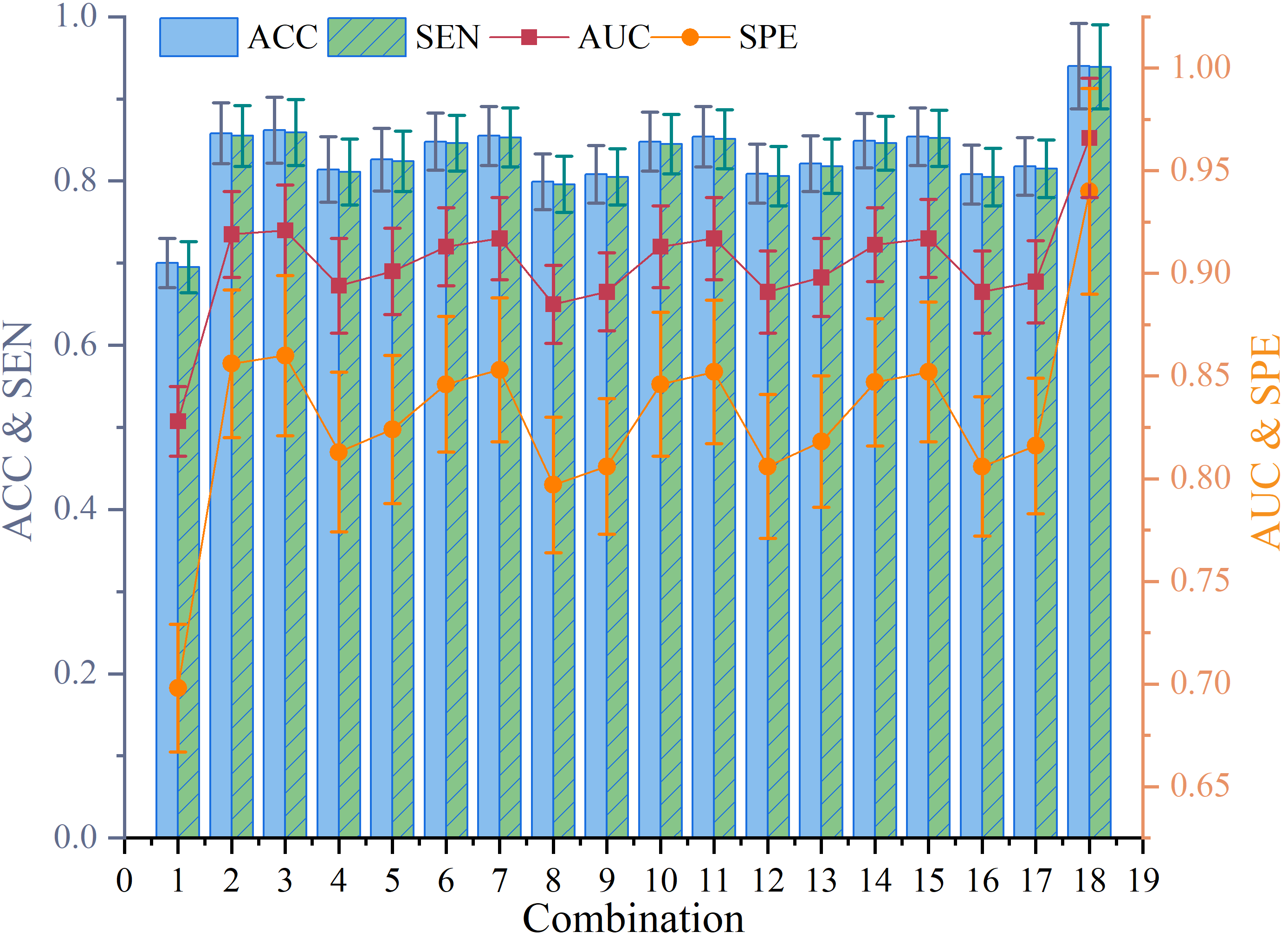}
\caption{The classification results using SVM with DE features to perform intra-subject recognition task of nine-category emotions on combination (1) \textbf{RAD}, combination (2)-(17) and combination (18) \textbf{GD} corresponding to Table \ref{tab:table3} No.1-18 respectively.}
\label{fig:fig12}
\end{figure}

\begin{table*}[h]
\centering
\caption{The results of competing models for categorizing the across-subject and intra-subjetc nine emotions on the original dataset as well as the generated data based on the combination No.2 of Table \ref{tab:table3}, where Raw\_data (\textbf{RAD}) denotes the original dataset, and \({\textbf{GD}_{\alpha1}}\) denotes data generated based on the generative path (Fig. \ref{fig:fig8}) of neutral, as well as negative and positive sentiment generation paths using (I) and (VI) in Fig. \ref{fig:fig9} respectively. CL denotes the classification task with cross or intra subject level.}
\begin{tabular}{c|c|c|cccc}
\hline
\textbf{Model} & \textbf{CL} & \textbf{Data} & \textbf{ACC}  & \textbf{AUC}  & \textbf{SEN}     & \textbf{SPE}     \\ \hline
\multirow{4}*{SVM\cite{hearst1998support} } & \multirow{2}*{Cross}       & \textbf{RAD}     & 0.430$\pm$0.051  & 0.675$\pm$0.029  & 0.421$\pm$0.051& 0.424$\pm$0.050    \\ 
               &  & \({\textbf{GD}_{\alpha1}}\) & \textbf{0.678$\pm$0.026} & \textbf{0.817$\pm$0.014} & \textbf{0.674$\pm$0.025} & \textbf{0.681$\pm$0.029} \\ \cline{3-7} 
               & \multirow{2}*{Intra} & \textbf{RAD}  & 0.700$\pm$0.030   & 0.828$\pm$0.017   & 0.695$\pm$0.031          & 0.698$\pm$0.031          \\  
               &   & \({\textbf{GD}_{\alpha1}}\)   & \textbf{0.940$\pm$0.052} & \textbf{0.966$\pm$0.029} & \textbf{0.939$\pm$0.051} & \textbf{0.940$\pm$0.050} \\ \hline
\multirow{4}*{RandomForest\cite{breiman2001random} }  & \multirow{2}*{Cross}& \textbf{RAD} & 0.381$\pm$0.049& 0.649$\pm$0.028  & 0.375$\pm$0.050 & 0.370$\pm$0.057   \\ 
               &   & \({\textbf{GD}_{\alpha1}}\)   & \textbf{0.818$\pm$0.049} & \textbf{0.887$\pm$0.053} & \textbf{0.815$\pm$0.047} & \textbf{0.818$\pm$0.056} \\ \cline{3-7} 
               & \multirow{2}*{Intra}& \textbf{RAD}  & 0.968$\pm$0.061  & 0.982$\pm$0.035   & 0.968$\pm$0.062          & 0.968$\pm$0.060          \\ 
               &  & \({\textbf{GD}_{\alpha1}}\)    & \textbf{0.973$\pm$0.055} & \textbf{0.985$\pm$0.031} & \textbf{0.973$\pm$0.055} & \textbf{0.973$\pm$0.054} \\ \hline
\multirow{4}*{Adaboost\cite{schapire2013explaining} }  & \multirow{2}*{Cross}  & \textbf{RAD} & 0.346$\pm$0.058 & \textbf{0.631$\pm$0.032}  & 0.343$\pm$0.056  & \textbf{0.331$\pm$0.059}      \\ 
               &  & \({\textbf{GD}_{\alpha1}}\)  & \textbf{0.430$\pm$0.008} &  0.499$\pm$0.001 & \textbf{0.413$\pm$0.008}  & 0.330$\pm$0.018 \\ \cline{3-7} 
               & \multirow{2}*{Intra} & \textbf{RAD} & \textbf{0.473$\pm$0.018} & \textbf{0.702$\pm$0.010}    & \textbf{0.470$\pm$0.018}    & \textbf{0.463$\pm$0.018}    \\ 
               &  & \({\textbf{GD}_{\alpha1}}\) & 0.429$\pm$0.021 & 0.497$\pm$0.005   & 0.412$\pm$0.018  & 0.326$\pm$0.027  \\ \hline
\multirow{4}*{Xgboost\cite{Chen:2016:XST:2939672.2939785} } & \multirow{2}*{Cross} & \textbf{RAD}& 0.374$\pm$0.046 & 0.646$\pm$0.026 & 0.371$\pm$0.047  & 0.363$\pm$0.052   \\ 
               & & \({\textbf{GD}_{\alpha1}}\) & \textbf{0.805$\pm$0.053} & \textbf{0.889$\pm$0.028} & \textbf{0.802$\pm$0.050} & \textbf{0.812$\pm$0.040} \\ \cline{3-7} 
               & \multirow{2}*{Intra}       & \textbf{RAD}   & 0.806$\pm$0.042          & 0.890$\pm$0.024  & 0.804$\pm$0.043          & 0.807$\pm$0.042          \\  
               && \({\textbf{GD}_{\alpha1}}\)   & \textbf{0.931$\pm$0.049} & \textbf{0.961$\pm$0.028} & \textbf{0.930$\pm$0.049} & \textbf{0.932$\pm$0.048} \\ \hline
\multirow{4}*{Naive Bayes}  & \multirow{2}*{Cross}  & \textbf{RAD}   & 0.313$\pm$0.033          & 0.613$\pm$0.019          & 0.311$\pm$0.034   & 0.358$\pm$0.053   \\ 
               & & \({\textbf{GD}_{\alpha1}}\) & \textbf{0.737$\pm$0.054} & \textbf{0.851$\pm$0.028} & \textbf{0.736$\pm$0.049} & \textbf{0.757$\pm$0.038} \\  \cline{3-7} 
               & \multirow{2}*{Intra}       & \textbf{RAD}      & 0.364$\pm$0.014          & 0.642$\pm$0.008          & 0.363$\pm$0.014    & 0.438$\pm$0.025          \\ 
               & & \({\textbf{GD}_{\alpha1}}\)   & \textbf{0.765$\pm$0.023} & \textbf{0.867$\pm$0.013} & \textbf{0.763$\pm$0.022} & \textbf{0.770$\pm$0.021} \\ \hline
\multirow{4}*{ResNet\cite{he2016deep}}   & \multirow{2}*{Cross}   & \textbf{RAD}      & 0.577$\pm$0.167     & 0.890$\pm$0.104   & 0.576$\pm$0.166   & 0.614$\pm$0.183  \\ 
               & & \({\textbf{GD}_{\alpha1}}\)   & \textbf{0.680$\pm$0.121} & \textbf{0.951$\pm$0.039} & \textbf{0.668$\pm$0.126} & \textbf{0.707$\pm$0.131} \\ \cline{3-7} 
               & \multirow{2}*{Intra}   & \textbf{RAD}   & 0.844$\pm$0.223          & 0.953$\pm$0.089          & 0.843$\pm$0.222          & 0.850$\pm$0.212          \\ 
               && \({\textbf{GD}_{\alpha1}}\)   & \textbf{0.862$\pm$0.165} & \textbf{0.978$\pm$0.044}  & \textbf{0.864$\pm$0.164} & \textbf{0.867$\pm$0.161}\\ \hline
\multirow{4}*{EfficientNet\cite{tan2019efficientnet}}   & \multirow{2}*{Cross}  & \textbf{RAD}  & 0.601$\pm$0.111 & 0.923$\pm$0.071 & 0.607$\pm$0.112& 0.664$\pm$0.133   \\  
               & & \({\textbf{GD}_{\alpha1}}\) & \textbf{0.708$\pm$0.104} & \textbf{0.961$\pm$0.029} & \textbf{0.697$\pm$0.108} & \textbf{0.723$\pm$0.113} \\ \cline{3-7} 
               & \multirow{2}*{Intra}   & \textbf{RAD}           & 0.812$\pm$0.183    & 0.960$\pm$0.072   & 0.818$\pm$0.112  & \textbf{0.842$\pm$0.153}          \\  
               &  & \({\textbf{GD}_{\alpha1}}\)  & \textbf{0.834$\pm$0.160} & \textbf{0.970$\pm$0.041} & \textbf{0.821$\pm$0.152} & 0.837$\pm$0.151 \\ \hline
\end{tabular}
\label{tab:table4}
\end{table*}

\subsection{Results of Emotion Recognition}
\label{subsec:4.6}
In addition, we implement signal generation for each emotion according to the generation path in Fig. \ref{fig:fig8} and Fig. \ref{fig:fig9}, and use SVM \cite{hearst1998support} with differential entropy (DE) features \cite{6695876} to perform recognition task of nine-category emotions on the original dataset and the generated data to verify whether the generated data conforms to the data distribution of the original dataset. Besides, other classical machine learning methods\cite{ breiman2001random}, \cite{schapire2013explaining}, \cite{Chen:2016:XST:2939672.2939785} combing DE features are also used for both cross-subject emotion recognition and intra-subject emotion recognition. In intra-subject emotion recognition, each video clip's EEG data is split 90\% for training and 10\% for testing, while in cross-subject emotion recognition, subjects are divided into 10 folds with a nine-fold training and one-fold testing split, repeated 10 times with classification performances averaged across 10 folds. Better yet, as illustrated in Tables \ref{tab:table2}, the data generated based on the signal generation paths for the nine emotions as described in Fig. \ref{fig:fig8} and Fig. \ref{fig:fig9} exhibits remarkable efficacy by five classical machine learning methods. Furthermore, for algorithm developers, they are more inclined to design deep learning modules to compete \cite{suk2012novel}, \cite{li2020quantum}, \cite{li2020perils}, \cite{qi2015rstfc}, \cite{ding2023lggnet}, \cite{zeng2015optimizing}. For this reason, we additionally use ResNet \cite{he2016deep} and EfficientNet \cite{tan2019efficientnet}, without DE features to perform the nine-category emotion recognition task. The results are also shown in Table \ref{tab:table2}, with remarkable improvement. These mean that our model can improve the quality of the dataset and greatly reduce data noise caused by some unexpected events during the collection process, and indicate that the EEG signals data produced by our generation path across nine emotions not only aligns with the actual data distribution but also demonstrates that, through the modeling of Hypotheses 1-3 (\ref{subsubsec:3.2.1}), these paths can effectively integrate emotions with corresponding EEG signal areas for latent alignment. Moreover, for downstream tasks, the performance of simply using traditional algorithms with feature engineering to recognize generated signals can match the recognition capabilities of complex neural network models, which suggests that the generated data will facilitate the development of downstream identification models and other applications. We also use spectral clustering \cite{von2007tutorial} to cluster the original data and generated data. Subsequently, we visualize the results using t-SNE \cite{van2008visualizing} as shown in Fig. \ref{fig:fig10}, which further proves the efficacy of our generation model.

Furthermore, we also paired the four signal generation paradigms of negative emotions with the four signal generation paradigms of positive emotions, that is, any paradigm of negative emotions was used to generate signals of other negative emotions, and the same operation was performed for positive emotions, while the signal generation method of neutral emotion remained unchanged, for a total of 16 combinations, as shown in Table \ref{tab:table3} No.2-17. We adopt the same scheme as in Table \ref{tab:table2} to perform recognition tasks of nine-category emotions on the generated data. The results for combination No.2 (Anger-Neutral-Amusement) are presented in Table \ref{tab:table4}, while the results for other categories are provided in the supplementary material.
At the same time, Fig. \ref{fig:fig11} and \ref{fig:fig12} successively show the performance of cross-subject and intra-subject recognition task of nine-category emotions based on combination (1) raw data (\textbf{RAD}), combination (2)-(17) and combination (18) Generated\_division (\textbf{GD}) responding to Table \ref{tab:table3} No.1-18 respectively, using SVM with DE features. Base on these results, we can conclude that under the condition of using the same recognition model, \({\textbf{GD}}\) is much more effective than \textbf{RAD}.

\section{Discussion and Conclusion}
\label{sec:5}
We first theoretically formulate the problem of generating dense-channel EEG signals from sparse channels by optimizing a set of cross-channel EEG generation tasks. Then, We introduce the novel unified framework YOAS for dense-channel EEG generation, leveraging prior information on electrode positioning from the International 10-20 system EEG placement, precise bias EEG data for each channel, and general across-channel EEG synthesis paradigms. Additionally, we validate three conjectures of generation pathways (depicted in Fig. \ref{fig:fig2}), contributing to the advancement of dense-channel EEG signal generation methodologies and pioneering frontiers in the field.

The proposed YOAS currently falls within the realm of data-driven statistical models, where threshold values for outlier detection are predominantly determined empirically. Future work may explore the adaptive detection of outliers in the Data Preparation stage of YOAS, drawing inspiration from metrics such as the Interquartile Range (IQR) \cite{wang2015novel}. Furthermore, parameters $L$ and $P$ are empirically set, and their correlation with physical mechanisms remains to be investigated in future research.
\\\hspace*{0.4cm}Our innovative approach revolutionizes the construction of dense-channel EEG systems by developing advanced artificial intelligence models, promising broader applications. By optimizing EEG systems to support more channels with fewer resources, we potentially significantly reduce the cost of high-density electrode arrays while maintaining performance. This breakthrough enhances EEG accessibility in healthcare, consumer electronics, and neuroscience.




\bibliographystyle{IEEEtran}
\bibliography{references}

\vfill
\end{document}